\documentclass[10pt,twocolumn,letterpaper]{article}

\usepackage[pagenumbers]{cvpr} 

\usepackage{indentfirst}
\usepackage{multirow}
\usepackage{makecell}
\usepackage{float}
\usepackage{caption}
\usepackage{graphicx}
\usepackage{subcaption}
\usepackage[accsupp]{axessibility}
\PassOptionsToPackage{bookmarks=false}{hyperref}

\definecolor{cvprblue}{rgb}{0.21,0.49,0.74}
\usepackage[pagebackref,breaklinks,colorlinks,allcolors=cvprblue]{hyperref}

\def\paperID{9285} 
\def\confName{CVPR}
\def\confYear{2025}

\title{PI-HMR: Towards Robust In-bed Temporal Human Shape Reconstruction with Contact Pressure Sensing}

\author{Ziyu Wu\footnotemark[1], Yufan Xiong\footnotemark[1], Mengting Niu, Fangting Xie, Quan Wan, Qijun Ying, Boyan Liu, Xiaohui Cai\footnotemark[2]\\
University of Science and Technology of China\\
}

\begin{document}

\crefname{section}{Sec.}{Secs.}
\Crefname{section}{Section}{Sections}
\Crefname{table}{Table}{Tables}
\crefname{table}{Tab.}{Tabs.}
\Crefname{figure}{Figure}{Figures}
\crefname{figure}{Fig.}{Figs.}

\setlength{\floatsep}{6pt plus 1pt minus 2pt}
\setlength{\textfloatsep}{4pt plus 1pt minus 2pt}
\setlength{\dbltextfloatsep}{4pt plus 1pt minus 2pt}
\setlength{\dblfloatsep}{4pt plus 1pt minus 2pt}
\setlength{\intextsep}{0pt}
\setlength{\abovecaptionskip}{3pt}
\setlength{\belowcaptionskip}{1pt}
\setlength{\parskip}{0pt}
\setlength{\abovedisplayskip}{0pt}
\setlength{\belowdisplayskip}{0pt}
\setlength\abovedisplayshortskip{0pt}
\setlength\belowdisplayshortskip{0pt}

\twocolumn[{
\renewcommand\twocolumn[1][]{#1}
\maketitle
\begin{center}
    \centering
    \vspace{-2.5em}
  \includegraphics[width=1.\linewidth]{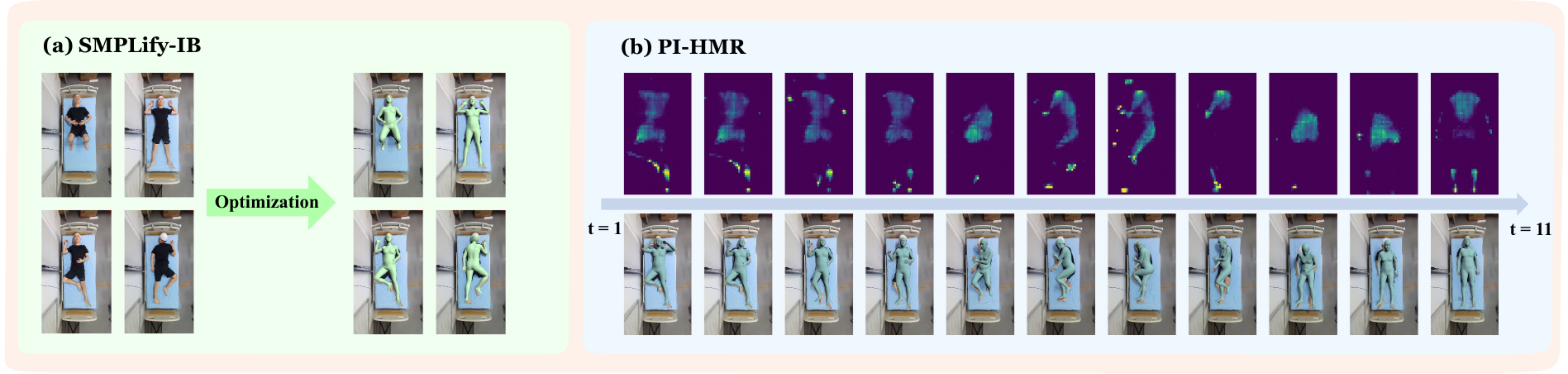}
  \captionof{figure}{We present a general framework for in-bed HPS tasks, containing a monocular optimization strategy to generate high-quality SMPL annotations in in-bed scenarios, SMPLify-IB; and a HPS network to predict in-bed motions from pressure sequence, PI-HMR.}
  \vspace{-0.5em}
  \label{fig: dataset_vis}
\end{center}
}]

\renewcommand{\thefootnote}{\fnsymbol{footnote}} 
\footnotetext[1]{These authors contributed equally to this work.} 
\footnotetext[2]{Corresponding authors.} 

\begin{abstract}
Long-term in-bed monitoring benefits automatic and real-time health management within healthcare, and the advancement of human shape reconstruction technologies further enhances the representation and visualization of users' activity patterns. However, existing technologies are primarily based on visual cues, facing serious challenges in non-light-of-sight and privacy-sensitive in-bed scenes. Pressure-sensing bedsheets offer a promising solution for real-time motion reconstruction. Yet, limited exploration in model designs and data have hindered its further development. To tackle these issues, we propose a general framework that bridges gaps in data annotation and model design. Firstly, we introduce SMPLify-IB, an optimization method that overcomes the depth ambiguity issue in top-view scenarios through gravity constraints, enabling generating high-quality 3D human shape annotations for in-bed datasets. Then we present PI-HMR, a temporal-based human shape estimator to regress meshes from pressure sequences. By integrating multi-scale feature fusion with high-pressure distribution and spatial position priors, PI-HMR outperforms SOTA methods with 17.01mm Mean-Per-Joint-Error decrease. This work provides a whole tool-chain to support the development of in-bed monitoring with pressure contact sensing.


\end{abstract}    
\section{Introduction}

Long-term and automatic in-bed monitoring draws increasing attention in recent years for the growing need in heathcare, such as sleep studies~\cite{chang2018sleepguard}, bedsore prevention~\cite{yousefi2011bed}, and detection of bed-exit and fall events~\cite{hoque2010monitoring}. The advancement of parameterized human representation~(\eg SMPL~\cite{loper2023smpl}) and human pose and shape estimation~(HPS) technologies further furnish technical underpinning for the reconstruction and visualization of patient motions, facilitating caregivers to comprehend patients' behavioral patterns in time. However, vision-based techniques, trained on in-lab or in-wild public datasets, fail in in-bed scenarios for more challenges are raised like poor illumination, occlusion by blankets, domain gaps with existing datasets~(\eg 3DPW~\cite{von2018recovering}), and privacy issues in both at-home or ICUs.

Our intuition lies in that tactile serves as a crucial medium for human perception of the surroundings. Especially for in-bed scenarios, lying postures prompt full engagement between humans and environment; simultaneously, this tactile perception also encompasses valuable information about their physiques. Reconstructing human motions from this tactile feedback might provide a privacy-preserving solution to automatic in-bed management for patients and elders. Thus, many efforts have been devoted to capturing the contact pressure with a pressure-sensing bedsheet, which integrates a pressure-sensitive sensor array and collects matrix-formatted pressure distribution~(named pressure images), and exploring potentials of full-body human reconstruction from these tactile sensors~\cite{clever20183d, clever2020bodies, tandon2024bodymap}. However, current methods are often constrained by model design, dataset diversity and label quality. The limitations can be categorized into three points: 

(1) \textbf{Lack of explorations on the pressure nature}. Despite both RGB and pressure images sharing similar structures, the meaning of each pixel differs significantly. For visual images, both foreground and background pixels are non-trivial, conveying texture and semantics. Nonetheless, with single-channel pressure data, regions lacking applied pressure are denoted as zeros, resulting in a dearth of semantic cues regarding the background. Furthermore, the relationship between pressure contours and human shapes introduces information ambiguity~\cite{tandon2024bodymap, yin2022multimodal} when some crucial joints do not directly interact with sensors. Previous research~\cite{clever2020bodies, tandon2024bodymap} attempted to estimate pressure based on the penetration depth of the human model and contact surfaces, thereby explicitly introducing pressure supervision. However, due to limitations in SMPL vertices granularity, sensor resolution, and tissue deformation, SMPL struggles to describe the contact mode with outsides, thus potentially impairing model performance. Consequently, hasty adoption of visual pipelines, without tailored design for pressure characteristics, might restrict model performance.

(2) \textbf{Limited data diversity}. Data diversity implicates models' generalization to unseen situations. For vision-based HPS tasks, the flourishing of HPS community is contributed by large-scale general~(\eg ImageNet~\cite{deng2009imagenet}) or task-specific~(\eg AMASS~\cite{mahmood2019amass}) datasets and mass of unlabeled data from Internet. However, as a human-centric and sensor-based task, in addition to the SLP~\cite{liu2022simultaneously} dataset that contains data from 102 individuals, most in-bed pressure datasets include fewer than 20 participants. Furthermore, the disparities of the sensor scale and performance across different studies, making it challenging to integrate these datasets, thus leading to poor performance to out-of-distribution users or motions. Therefore, how to learn priors across datasets and modalities is of paramount significance.



(3) \textbf{Limited 3D label quality}. One main factor limiting the data diversity is the challenge of acquiring accurate 3D labels, especially for an in-bed setting. Currently, only SLP~\cite{liu2022simultaneously} and TIP~\cite{wu2024seeing} datasets offer both SMPL pseudo-ground truth~(p-GTs) and RGB images, with annotations in TIP being seriously doubted by depth ambiguity and penetrations due to monocular SMPLify-based optimization~(in~\cref{fig: dataset_vis}). Limited label quality might lead the model to misinterpret pressure cues, thus calling for a low-cost and accurate label annotation approach for in-bed scenes.

To tackle aforesaid disparities, in this work, we present a general framework bridging from annotations, model design and evaluation for pressure-based in-bed HPS tasks. Concretely, we firstly present PI-HMR, a pressure-based in-bed human shape estimation network to predict human motions from pressure sequences, as a preliminary exploration to utilize pressure characteristics. Our core philosophy falls that both joint positions and contours of high-pressure areas are essential to sense pressure distribution and its variation patterns from the redundant zero-value backgrounds. Thus, we achieve this by explicitly introducing these semantic cues, compelling the model to focus on core regions by feature sampling. Furthermore, considering that the sensing mattress is often fixed in the environment, we leverage these positional priors and feed them into the model to learn the spatial relationship between humans and sensors. Experiments show that PI-HMR brings 17.01mm MPJPE decrease compared to PI-Mesh~\cite{wu2024seeing} and outperforms vision-based temporal SOTA architecture TCMR~\cite{choi2021beyond}~(re-trained on pressure images) with 4.91mm MPJPE improvement.

Moreover, to further expand prior distribution within limited pressure datasets, we realize (1) a Knowledge Distillation~(KD)~\cite{hinton2015distilling} framework to pre-train PI-HMR's encoder with RGB-based SOTA method CLIFF~\cite{li2022cliff}, to facilitate cross-modal body and motion priors transfer; and (2) a pre-trained VQ-VAE~\cite{van2017neural} network as in-bed motion priors in a unsupervised Test-Time Optimization to alleviate information ambiguity. Experiments show that both modules bring 2.33mm and 1.7mm MPJPE decrease, respectively.

Finally, for a low-cost but efficient label annotation method tailored for in-bed scenes, we present a monocular optimization approach, SMPLify-IB. It incorporates a gravity-constraint term to address depth ambiguity issues in in-bed scenes, and integrates a potential-based penalty term with a lightweight self-contact detection module to alleviate limb penetrations. We re-generated 3D p-GTs in the TIP~\cite{wu2024seeing} dataset and results show that SMPLify-IB not only provides higher-quality annotations but also mitigates implausible limb lifts. This suggests the feasibility of addressing depth ambiguity issues with physical constraints in specific scenarios. Besides, results prove that our detection module is 53.9 times faster than SMPLify-XMC~\cite{muller2021self} while achieving 98.32\% detection accuracy. 

We highlight our key contributions: (1) a general framework for pressure-based in-bed human shape estimation task, spanning from label generation to algorithm design. (2) PI-HMR, a temporal network to directly predict 3D meshes from in-bed pressure image sequences and outperforms both SOTA pressure-based and vision-field architectures. (3) SMPLify-IB, a gravity-based optimization technique to generate reliable SMPL p-GTs for monocular in-bed scenes. Based on SMPLify-IB, we re-generate 3D annotations for a public dataset, TIP, providing higher-quality SMPL p-GTs and mitigating implausible limb lifts due to depth ambiguity. (4) We explore the feasibility of prior expansion with knowledge distillation and TTO strategy.  

\begin{figure}[t]
  \centering
  \includegraphics[width=\linewidth]{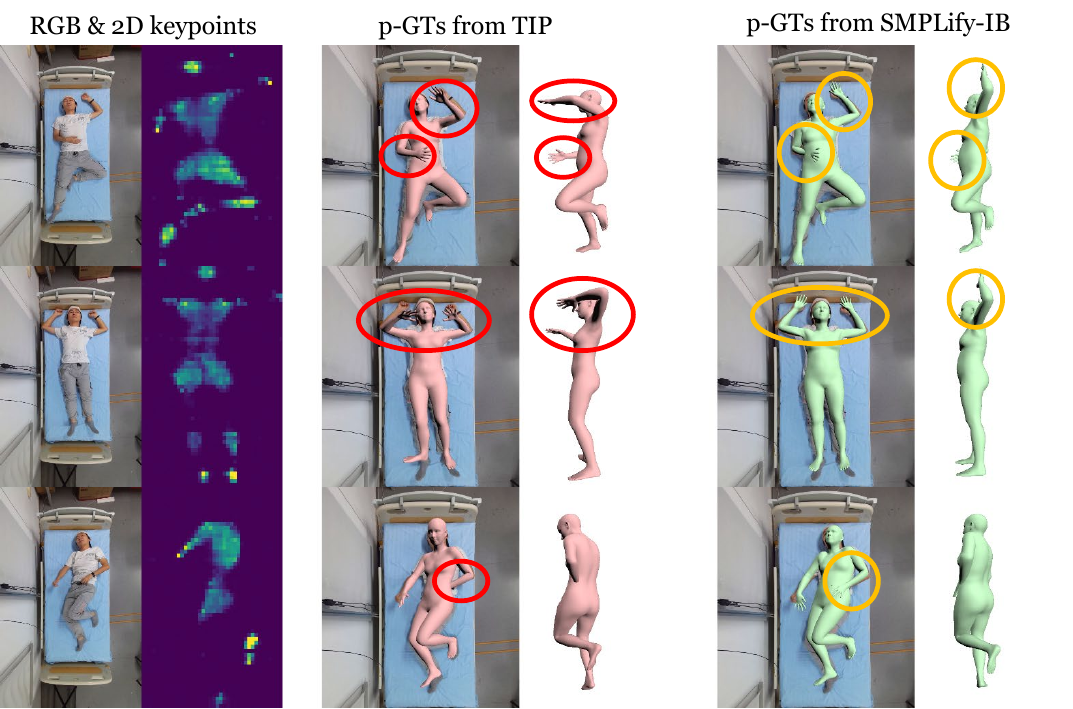}
  \caption{A glimpse of TIP dataset, with p-GTs from TIP and our SMPLify-IB. we highlight its drawbacks with red ellipses and our refinements in yellow ones.}
  \label{fig: dataset_vis} 
\end{figure} 
\section{Related Work}
\textbf{Regression for HPS.} Recent years have witnessed tremendous advances in vision-based human shape reconstruction approaches from images~\cite{kanazawa2018end, kolotouros2019learning, kocabas2021spec, zhang2021pymaf, kocabas2021pare, li2022cliff, wang2023refit, shimada2023decaf, goel2023humans, dwivedi2024tokenhmr, shin2024wham, song2024posturehmr} based on the parametric human body model~(\ie, SMPL~\cite{loper2023smpl}). Meanwhile, several works take video clips as input to exploit the temporal cues~\cite{kanazawa2019learning, kocabas2020vibe, choi2021beyond, wei2022capturing, shen2023global, you2023co}, utilizing the temporal context to improve the smoothness. 



We mainly focus on HPS from contact pressure sensing. Unlike visual information, the representation pattern of contact pressure data is influenced by its perceptual medium, thus necessitating a corresponding alteration in algorithm design. Typical sensing devices, combined with HPS algorithms,~(\eg, carpets~\cite{luo2021intelligent, chen2024cavatar}, clothes~\cite{zhou2023mocapose, zhang2024learn}, bedsheets or mattress~\cite{liu2022simultaneously, wu2024seeing, clever2020bodies, tandon2024bodymap}, and shoes~\cite{zhang2024mmvp, van2024diffusionposer}), are applied as a major modality or supplements to help generate robust body predictions in pre-defined scenes or tasks. Nevertheless, the process strategy of pressure data leans on vision pipelines, lacking a thorough contemplation of its inherent nature. 


\textbf{Optimization for HPS.} Optimization-based methods typically fit the SMPL parameters to image cues~\cite{bogo2016keep, pavlakos2019expressive}~(\eg detected 2D joints~\cite{cao2017realtime, xu2022vitpose}), combined with data and prior terms. Follow-up studies further introduced supplement supervisions, including, but not limited to temporal consistency~\cite{arnab2019exploiting}, environment~\cite{kaufmann2023emdb}, human-human/scene contact~\cite{hassan2019resolving, muller2024generative, huang2024intercap}, self-contact~\cite{muller2021self} and large language models~(LLMs)~\cite{subramanian2024pose} to regularize motions in specific context. Besides, in recent years, efforts have emerged to integrate both optimization and regression methods as a cheap but effective annotation technique to produce pseudo-labels for visual datasets~\cite{wu2024seeing, zhang2024mmvp, huang2024intercap}, especially for monocular data from online images and videos~\cite{joo2021exemplar, muller2021self, lin2023one, yi2023generating}. 




\textbf{In-bed human pose and shape estimation. } 
Compared with other human-related tasks, in-bed HPS faces more serious challenges from data quality and privacy issues. Thus, efforts are devoted to pursuing environmental sensors for in such a non-light-of-sight~(NLOS) scenes, such as infrared camera~\cite{liu2019seeing, liu2019bed}, depth camera~\cite{grimm2012markerless, achilles2016patient, clever2020bodies}, pressure-sensing mattresses~\cite{clever20183d, clever2020bodies, davoodnia2023human, wu2024seeing}. Specifically for pressure-based approaches, \citet{clever2020bodies} conducted pioneering studies by involving pressure estimation to reconstruct in-bed shapes from a single pressure image~\cite{clever2020bodies}. \citet{wu2024seeing} collected a three-modality in-bed dataset TIP, and employed a VIBE-based network to predict in-bed motions from pressure sequences. 
\citet{yin2022multimodal} proposed a pyramid scheme to infer in-bed shapes from aligned depth, LWIR, RGB, and pressure images, and \citet{tandon2024bodymap} improves accuracy on SLP~\cite{liu2022simultaneously} with depth and pressure modalities by integrating a pressure prediction module as auxiliary supervision. 





\section{Dataset and Label Enhancement}
\subsection{Data Overview}
We select TIP~\cite{wu2024seeing} as our evaluation dataset because, to our knowledge, it is the sole dataset containing both temporal in-bed pressure images and SMPL annotations. TIP is an in-bed posture dataset that contains over 152K synchronously-collected three-modal images~(RGB, depth, and pressure) from 9 subjects, with matched 2D keypoint and 3D SMPL annotations. We present a glimpse visualization in~\cref{fig: dataset_vis}. The SMPL annotations are generated by a SMPLify-like approach. However, we notice severe depth ambiguity~(\eg, mistaken limb lifts) and self-penetration in their p-GTs~(marked in~\cref{fig: dataset_vis}), which are common issues for monocular optimization. Considering that reliable labels are crucial for the robustness of algorithms, we presented a general optimization approach that utilizes physical constraints to generate accurate SMPL p-GTs for in-bed scenes, named SMPLify-IB, and re-generated annotations for the whole dataset. Compared with raw annotations, we have significantly enhanced the rationality of the labels~(shown in~\cref{fig: dataset_vis}). More results will be presented in~\cref{subsec:exp_re_SIB}.

\subsection{SMPLify-IB: Generate reliable p-GTs for TIP}
SMPLify-IB contains two core alterations compared with traditional approaches: a gravity-based constraints to penalize implausible limb lift due to depth ambiguity, and a lightweight penetration detection algorithm with a potential-based loss term to penalize self-penetration. We briefly summarize our efforts as follows, and more details are given in the Sup. Mat..

\subsubsection{Gravity Constraint}
To tackle the implausible limb lifts, our rationale lies in the observations that when a person lies in bed, it should stay relaxed. Conversely, when limbs are intentionally lifted, a torque is generated at the shoulders or hips, thus resulting in discomfort. Such a conflict inspires us that when a person is motionless, all limbs should receive support to avoid an "uncomfortable" posture. Based on such an intuition, we propose a zero-velocity detection algorithm to detect implausible limb suspensions caused by depth ambiguity and exert gravity constraints to push them into contact with the bed plane or other body parts for support. Specifically, we use velocities of 2D keypoint ground-truths to calibrate limb status. For those velocities exceeding a pre-defined threshold,  we consider them to be in normal movement states; for limbs raised but nearly static, we annotate them as miscalculations from depth ambiguity and punish their distance to the bed plane. The loss term is as follows:
\begin{equation}
\small
\mathcal{L}_{g}=\sum_{i}^T \sum_{\substack{j \in G_J \\
z(i)_j>0}} \mathbb{I}\left(\operatorname{v}(i)_j<t h r e_{v}\right) e^{w_j \cdot z(i)_j} \\
\end{equation}
where $G_J$ is the set of gravity-constrained limb joints including hands, elbows, knees, and ankles, $z(i)_j$ is the signed distance of joint $j$ in timestamp $i$ to the bed plane, $v(i)_j$ is its velocity, $thre_{v}$ is the velocity threshold, $\mathbb{I}$ is the indicator function, and $\omega_j$ is the hyperparameter. 

\subsubsection{Potential-based self-penetration Constraint}
In order to reduce complexity, we only penalize the distance between lifted limbs and the bed plane in gravity loss $L_g$, which might further exacerbate self-penetration. Thus, the other main goal of SMPLify-IB is to punish severe self-intersection while encouraging plausible self-contact. Given that the \textit{Self-Contact} approach in SMPLify-XMC~\cite{muller2021self} is slow for large-dataset annotation
, we propose a lightweight self-contact supervision that includes two main parts, lightweight self-penetration detection and potential-based self-penetration penalty modules.

\begin{figure}[t]
  \centering
  \includegraphics[width=\linewidth]{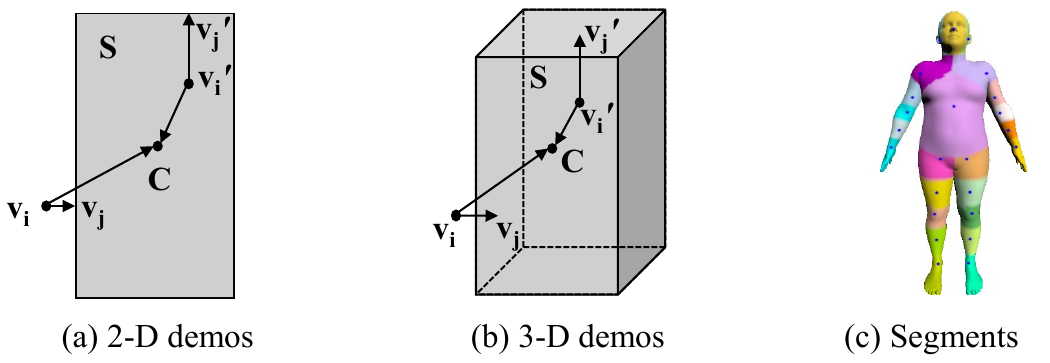}
  \caption{(a) and (b): demos of our detection algorithm. $S$ is the segment, $C$ is its segment center. $v_i$ are vertices that need to be checked for penetration with $S$, and $v_j$ are the vertices from $S$ that are closest to $v_i$, respectively. When $\overrightarrow{v_iv_j} \cdot \overrightarrow{v_iC} < 0$, $v_i$ is in penetration, and vice versa. (c) is our segment.} 
  \label{fig: dataset_pen_demo}
\end{figure}

\begin{figure*}[htbp]
  \centering
  
  \includegraphics[width=\linewidth]{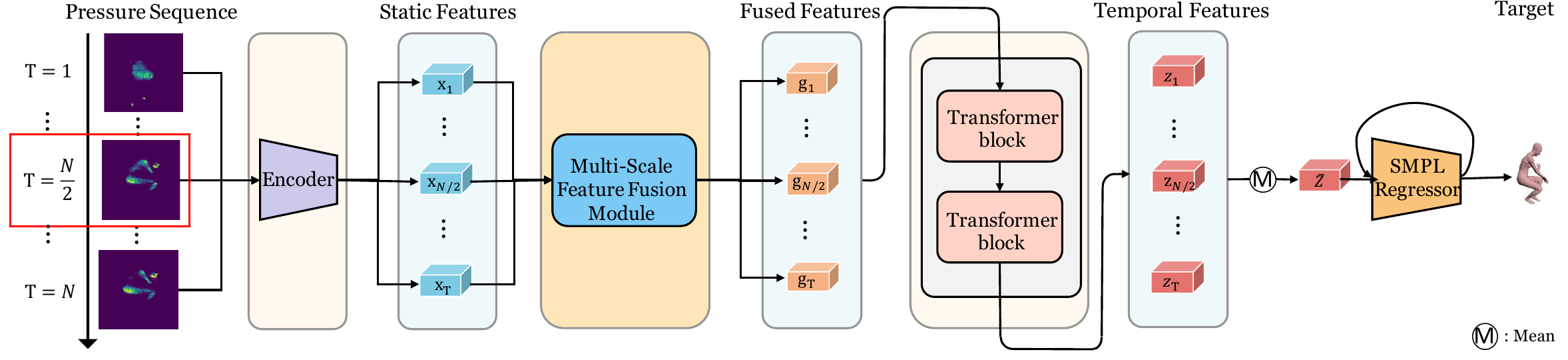}
  \caption{\textbf{An overview of PI-HMR.} PI-HMR outputs the midframe's SMPL predictions of the whole sequence.
  } 
  \label{fig: pihmr_architecture}
\end{figure*}

\textbf{Lightweight Detection.} In SMPLify~\cite{bogo2016keep}, authors used capsules to approximate human parts and calculate cross-part penetration. Although it's a coarse-grained limb representation, we notice that in such a capsule, the angle formed by the capsule center, penetrating vertex, and its closest vertex on the capsule-wall is likely to be obtuse. Following the observation, instead of calculating all solid angles between 6890 SMPL vertices and 13776 triangles in \textit{Winding Numbers}~\cite{jacobson2013robust} applied by SMPLify-XMC~\cite{muller2021self}, we make an approximation that the SMPL vertices could be viewed as an aggregation of multiple convex, uniform, and encapsulated segments~(shown in~\cref{fig: dataset_pen_demo}(c)), thus facilitating us to judge penetrations by spatial relations between vertices and segment centers. Specifically, assuming that a posed SMPL model could be represented by $K$ non-intersecting and convex vertex sets $\{S_1, ..., S_K\}$ and their segment center set $\{c_1, ..., c_K\}$. For any vertex $v_i$ from segment $S_i$, to determine whether it intersects with $S_j$, we firstly calculate its nearest vertex in $S_j$~(noted as $v_j$), and then judge whether intersection occurs for vertex $v_i$ by the sign of dot product $\overrightarrow{v_iv_j} \cdot \overrightarrow{v_ic_j}$~($\overrightarrow{v_iv_j} \cdot \overrightarrow{v_ic_j} < 0$ means $v_i$ is inside the segment $S_j$, and vice versa). We provide intuitively demos in~\cref{fig: dataset_pen_demo}. 

To construct approximately-convex segments, we pre-define 24 segment centers~(including 16 SMPL joints and 8 virtual joints in joint-sparse limbs like arms and legs to ensure uniformity), and employ a clustering algorithm to determine the assignment of SMPL vertices to segments. Finally, 24 segments are generated and visualized in~\cref{fig: dataset_pen_demo}(c).

\textbf{Potential-based constraints.} 
Beside commonly-used point-wise contact term~(noted as $\mathcal{L}_{p\_con}$) and penetration penalty~(noted as $\mathcal{L}_{p\_isect}$) with Signal Distance Field~(SDF) as specified in SMPLify-XMC, we notice that SMPL vertices are spatially influenced by their closest joints. Thus, we could directly penalize the distance between centers of two intersecting segments, to push these segments moving away. For two intersecting segments $S_i$ and $S_j$, and their centers $c_i$ and $c_j$, we denote the penalty as:
\begin{equation}
\small
    \mathcal{L}_{push} = |\mathbb{D}(S_i, S_j)| \exp (-\lambda_{push} ||c_i - c_j||)
\end{equation}
where $\mathbb{D}$ is the detection algorithm and $|\mathbb{D}(S_i, S_j)|$ means the number of intersecting vertices. Similarly, we use the same version to represent the self-contact term to encourage those close but non-intersected segments to contact:
\begin{equation}
\small
    \mathcal{L}_{pull} = -|\mathbb{C}(S_i, S_j)| \exp (-\lambda_{pull} ||c_i - c_j||)
\end{equation}
where $\mathbb{C}$ is the contact detection algorithm~(\ie, SDF of two vertices between 0 - 0.02m). Both terms are constrained by set scales and center distances, acting like repulsive forces between clusters, thus named potential constraints.

Finally, we could get the whole penetration term $L_{sc}$. 
\begin{equation}
\small
    \mathcal{L}_{sc} = \mathcal{L}_{p\_con} + \mathcal{L}_{p\_isect} + \mathcal{L}_{push} +  \mathcal{L}_{pull}
\label{eq:cp_term}
\end{equation}
\subsubsection{SMPLify-IB}
Finally, we present SMPLify-IB, a two-stage optimization method for SMPL p-GTs from monocular images. In the first stage, we use the CLIFF predictions as initialization and jointly optimize the shape parameter $\beta$, translation parameter $t$, and pose parameter $\theta$. After that, we use the mean shape parameters of all frames from the same subject as its shape ground truths. In the second stage, we freeze $\beta$ and only optimize $t$ and $\theta$. Both stages share the same objective functions~\cref{eq:oj_func}, exhibited as follows:
\begin{equation}
\small
\begin{split}
    \mathcal{L}_{IB} = \lambda_J \mathcal{L}_J + \lambda_p \mathcal{L}_p + \lambda_{sm} \mathcal{L}_{sm} + \lambda_{cons} \mathcal{L}_{cons}\\ 
    + \lambda_{bc} \mathcal{L}_{bc} + \lambda_{g} \mathcal{L}_{g} + \lambda_{sc} \mathcal{L}_{sc}
\end{split}
\label{eq:oj_func}
\end{equation}
Besides the gravity loss $\mathcal{L}_g$ and self-penetration term $\mathcal{L}_{sc}$, $\mathcal{L}_J$ and $\mathcal{L}_p$ denotes the re-projection term and prior term, as specified in~\cite{bogo2016keep}; $\mathcal{L}_{sm}$ is the smooth term and $\mathcal{L}_{cons}$ is the consistency loss that penalizes the differences between the overlapped part of adjacent batches; and $\mathcal{L}_{bc}$ is the human-bed penetration loss, which is the same as~\cite{wu2024seeing}.


\section{Method}
\subsection{PI-HMR}
Our motivation is to utilize pressure data nature. So our efforts fall into three stages: alleviating the dataset bottleneck and learning cross-dataset human and motion priors in the pre-training stage; pressure-based PI-HMR's design; and learning user's habits to overcome information ambiguity in the TTO. Thus, the data flow includes: (1) pre-train: KD-based pre-training with the training set; (2) train: train the PI-HMR and VQ-VAE with the training set; (3) test: test with PI-HMR on the test set and improve the estimates with the TTO strategy. \cref{fig: pihmr_architecture} shows the framework of PI-HMR. The details of each module will be elaborated as follows:

\subsubsection{Overall Pipeline of PI-HMR}
Given an input pressure image sequence $V=\{I_i \in \mathbb{R}^{H\times W}\}^T_{t=1}$ with $T$ frames, PI-HMR outputs the SMPL predictions of the mid-frame by a three-stage feature extraction and fusion modules. Following~\cite{kocabas2020vibe, choi2021beyond, wei2022capturing}, we first use ResNet50 to extract the static feature of each frame to form a static representation sequence $X = \{x_t \in \mathbb{R}^{2048 \times H_1 \times W_1}\}^T_{t=1}$. The extracted $X$ is then fed into our Multi-scale Feature Fusion module~(MFF) to generate the fusion feature sequences $G = \{g_t\}^T_{t=1}$, with two-layer Transformer blocks behind to learn their long-term temporal dependencies and yield the temporal feature sequence $Z = \{z_t\}^T_{t=1}$. Finally, We use the mean feature of $Z$ as the integrated feature representation of the mid-frame and produce final estimations with an IEF SMPL regressor~\cite{kanazawa2018end}.   

\subsubsection{Multi-Scale Feature Fusion Module}
\begin{figure}[tbp]
  \centering
  \includegraphics[width=\linewidth]{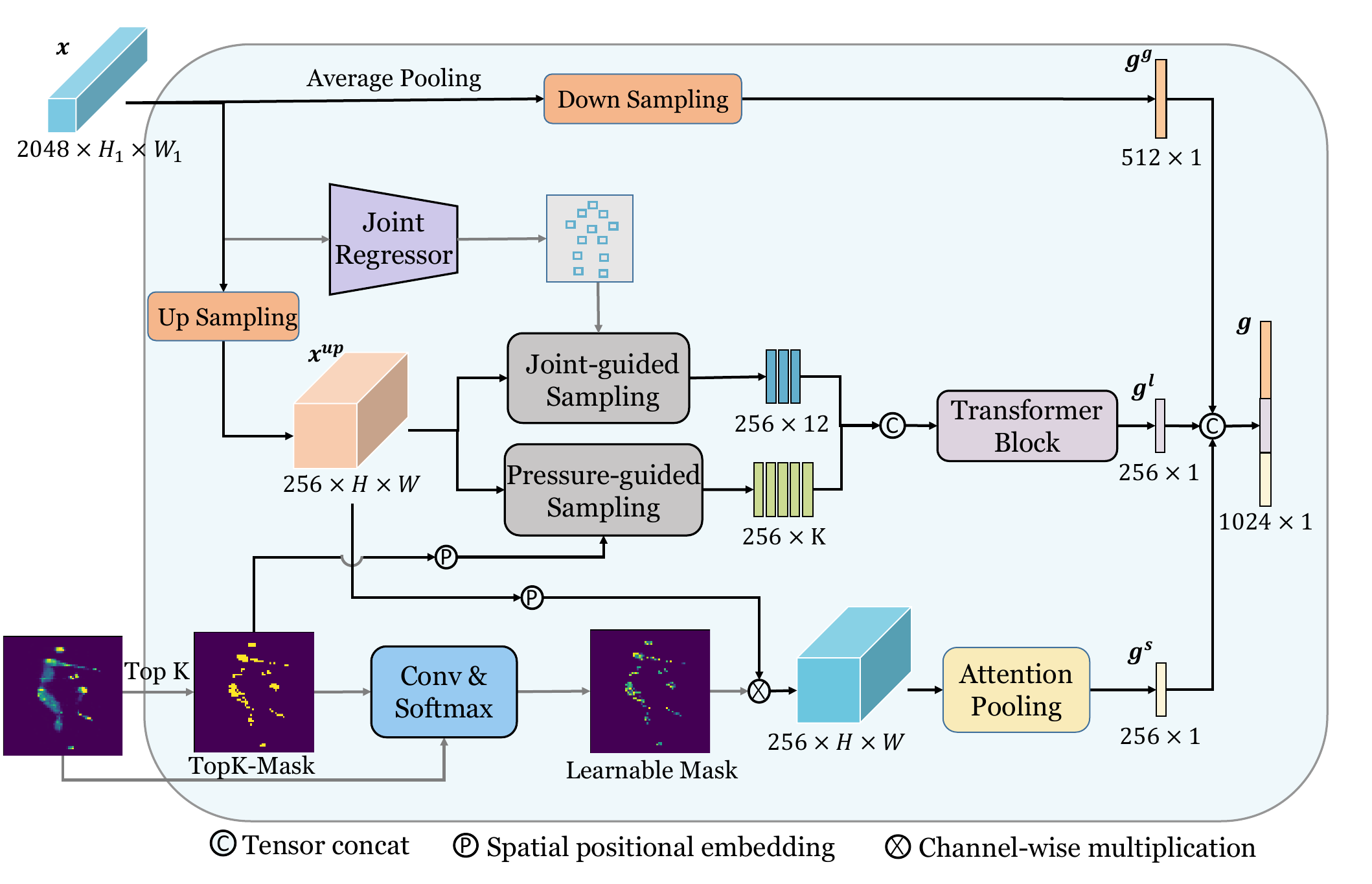}
  \caption{\textbf{Framework of our multi-scale feature fusion module.}} 
  \label{fig: sec7_pimesh_structure}
\end{figure}

To exploit the characteristics of pressure images, our core insight lies in that both large-pressure regions and human joint projections are essential for model learning: large-pressure regions represent the primary contact areas between humans and environments, directly reflecting user's posture and movement tendencies; 2D joint positions, always accompanied by inherent information ambiguity, serve to assist the model in learning the local pressure distribution pattern between small and large pressure zones. Following the insight, we present the Multi-scale Feature Fusion module~(MFF), shown in~\cref{fig: sec7_pimesh_structure}. MFF extracts multi-scale features from the static feature $x_i$ with the supervision of high-pressure masks and human joints, and generates the fusion feature $g_i$ for the next-stage temporal encoder. Before delving into MFF, we first introduce our positional encoding and high-pressure sampling strategy.

\textbf{Spatial Position Embedding.} We introduce a novel position embedding approach to fuse spatial priors into model learning. Compared with visual pixels, we could acquire the position of each sensing unit and their spatial relationships, given that the sensors remain fixed during data collection. Specifically, for a sensing unit located in pixel $(i, j)$ of a pressure image, we could get its position representation $[i, j, i \cdot d_h, j \cdot d_w]$, with $d_h$, $d_w$ being the sensor intervals along x-axis and y-axis~($d_h=0.0311m$ and $d_w=0.0195m$ in TIP). The first two values mean its position within image, while the latter ones denote the position in the world coordinate system (with its origin at the top-left pixel position of the pressure image). The representation is then transformed into spatial tokens $P \in \mathbb{R}^{256}$ using a linear layer. During the training, we could generate the spatial position map for the whole pressure image, noted as $P_i \in \mathbb{R}^{256 \times H \times W}$.

\textbf{TopK-Mask and Learnable Mask.} We employ a Top-K selection algorithm to generate  high-pressure 0-1 masks for each pressure image~(elements larger than K-largest value is set as 1). The mask, noted as $H^K$, will be fed into MFF as contour priors. Besides, we incorporate a learnable mask $H^{LK}$ into our model, utilizing the initial pressure input $I_i$ and the TopK-Mask matrix $H_i^K$ to learn an attention distribution that evaluates the contribution of features in the feature map. The learnable mask is computed as:
\begin{equation} 
    H^{LK}_i = \text{Softmax}(\text{Conv}([I_i \odot H_i^K, H^K_i]))
\end{equation}
where $\odot$ is the Hadamard product. The product result will be stacked with the TopK-mask and fed into a 1-layer convolution layer and Softmax layer to generate the attention matrix $H^{LK}_i \in \mathbb{R}^{H \times W}$. We aim to explicitly integrate these pressure distributions to enhance learnable masks' quality. The K is set as 128 in PI-HMR, and we also conduct ablations to discuss the selection of K in~\cref{tab: ablations for PI-HMR}.

\textbf{Auxiliary Joint Regressor.} We use an auxiliary joint regressor to provide 2D joints for the multi-scale feature extraction~(shown in~\cref{fig: sec7_pimesh_structure}). The regressor takes the static feature $x_i$ as input and returns the 2D positions of 12 joints in the pressure image, noted as $J_i^{2D}$. The 2D regressor will be trained in conjunction with the entire model.

\textbf{Multi-Scale Feature Fusion.}
We extract the global feature $g_i^g$, local feature $g_i^l$, and sampling feature $g_i^s$ from the static feature $x_i$, without replying on the temporal consistency. Firstly for global feature, we apply average pooling and downsampling to the static features $x_i \in \mathbb{R}^{2048 \times H_1 \times W_1}$ to generate global representation $g_i^g \in \mathbb{R}^{512}$. 

Subsequently, we perform dimension-upsampling on $x_i$ to obtain upsampled feature $x_i^{up} \in \mathbb{R}^{256 \times H \times W}$ that aligned with the initial pressure input scale, facilitating us to apply spatial position embedding and feature sampling. For local features, we add $x_i^{up}$ to the spatial position map $P_i$ we have learned, multiply it point-wise with the Learnable Mask $H^{LK}_i$, and then subject it to AttentionPooling to derive the local features $g_i^l \in \mathbb{R}^{256}$. 

As for the sampling features, we employ a feature sampling process on $x_i^{up}$ based on the pre-obtained TopK-Masks and 12 2D keypoint positions obtained from a auxiliary 2D keypoint regressor and get a medium feature $g_i^{mid} \in \mathbb{R}^{(K + 12) \times 256}$. After the same spatial position embedding, the medium feature will be input into a 1-layer Transformer layer to learn its spatial semantics, with the mean of the results serving as the sampling feature $g_i^s \in \mathbb{R}^{256}$.

Finally we get the fusion feature $ g_i \in \mathbb{R}^{1024}$ by concatenating aforesaid global, local, and sampling features.  

\subsubsection{Training Strategy}

The overall loss function can be expressed as follows:
\begin{equation} \label{eq: overall_func}
\small
   \mathcal{L}_{pi} = \lambda_{\text{SMPL}} \mathcal{L}_{\text{SMPL}} + \lambda_{3D} \mathcal{L}_{3D} + \lambda_{2D} \mathcal{L}_{2D} 
\end{equation}
where $\mathcal{L}_{\text{SMPL}}$ and $\mathcal{L}_{3D}$ presents the deviations between the estimated SMPL parameters and 3d joints with GTs,  and $\mathcal{L}_{2D}$ minimize errors in 2D joints for the auxiliary regressor.

\subsection{Encoder pre-train by cross-modal KD}
We employ a cross-modal KD framework to pretrain our PI-HMR's feature encoder, aiming at learning motion and shape priors from vision-based methods on paired pressure-RGB images. Specifically, we implement a HMR~\cite{kanazawa2018end} architecture as the student network $\mathcal{F}_S$~(with a ResNet50 as encoder and a IEF~\cite{kanazawa2018end} SMPL regressor), and choose CLIFF~(ResNet50)~\cite{li2022cliff} as the teacher model $\mathcal{F}_T$~(a HMR-based network). During pre-training, we apply extra feature-based and response-based KD~\cite{gou2021knowledge} to realize fine-grained knowledge transfer. Given input pressure-RGB-label groups~$(I_P, I_R, y)$, and 4 pairs of hidden feature maps from $\mathcal{F}_T$ and $\mathcal{F}_S$~(ResNet50 has 4 residual blocks, so we extract the feature maps after each residual block), i.e., $M_T$ from $\mathcal{F}_T$ and $M_S$ from $\mathcal{F}_S$, the loss function is:
\begin{equation}
\small
\begin{split}
    L_{KD} = \lambda_{kd}^y L_{pi}(\mathcal{F}_S(I_P), y) + \lambda_{kd}^T L_{pi}(\mathcal{F}_S(I_P), \mathcal{F}_T(I_R)) \\
    + \lambda_{kd}^F \sum_{i=1}^4||M_S^i - M_T^i|| 
\end{split}
\end{equation}
where $L_{pi}$ is the same as~\cref{eq: overall_func}, and $\lambda$ is the hyperparamter. After training and convergence, the ResNet50 encoder from $\mathcal{F}_S$ will be adopted as PI-HMR's pre-trained static encoder and finetuned in the following training process.

\setlength{\aboverulesep}{0pt}
\setlength{\belowrulesep}{0pt}
\begin{table*}[t] 
\small
\centering
\begin{tabular}{l|c|c|cccc}
\toprule[2pt]
Method       & Input  & Modalities      & MPJPE  & PA-MPJPE & MPVE  & ACC-ERR \\
\hline
HMR~\cite{kanazawa2018end} & \multirow{3}{*}{single} & \multirow{9}{*}{Pressure} & 75.06 & 57.97 &  89.11 &   31.52  \\
HMR-KD       &            &         &   66.30    &          52.41      &  83.01        &     24.41      \\
BodyMap-WS~\cite{tandon2024bodymap}   &           &        &   71.48    &     \textbf{40.91}      &   80.08       &   27.98     \\ \cline{1-2} \cline{4-7}
TCMR~\cite{choi2021beyond}         & \multirow{7}{*}{sequence}     &     & 64.37 &   46.76      &     74.66         &       20.12    \\
MPS-NET~\cite{wei2022capturing}      &               &         &         160.59    &    112.12     &     187.13        &      28.73         \\
PI-Mesh~\cite{wu2024seeing}      &        &         &    76.47   &     54.65     &   90.54       &    21.86  \\
 \cline{1-1} \cline{4-7}
PI-HMR (ours)       &         &      & 59.46         &        44.53      &      69.92    &       \textbf{9.12}     \\
PI-HMR + KD (ours)       &        &     &   \underline{57.13}  & 42.98         &  \underline{67.22}    &        9.84     \\
PI-HMR + TTO (ours)       &       &       &    57.76        &     43.31    &   67.76   &         \underline{9.83}    \\
PI-HMR + KD + TTO (ours) &        &       &      \textbf{55.50}     &    \underline{41.81}     &   \textbf{65.15}       &   9.96        \\
\bottomrule[2pt]
\end{tabular}
\caption{\textbf{Overall results of PI-HMR with SOTA methods}} \label{tab: overall_results}
\end{table*}
\begin{figure*}[t] \vspace{-0.1cm}
  \centering
  \includegraphics[width=\linewidth]{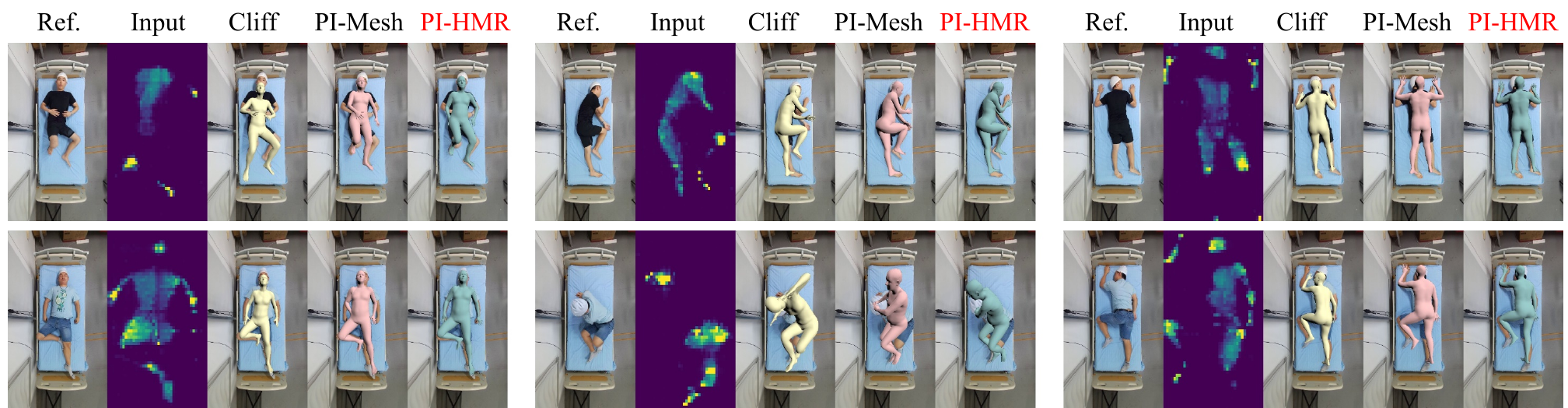}
  \caption{\textbf{Qualitative visualization for PI-HMR.} PI-HMR and PI-Mesh's results are generated by pressure images, while CLIFF's outputs are generated by RGB images for cross-modal comparison. Predictions are rendered on RGB images for comparison convenience}
  \label{fig: overall_vis}
\end{figure*} 

\subsection{Test-Time Optimization}
We also explore a TTO routine to further enhance prediction quality of PI-HMR. Considering that there hasn't been a general 2D keypoint regressor for pressure images, we are inclined toward seeking an unsupervised, prior-based optimization strategy. We notice that humans exhibit similar movement patterns across various postural states~(\eg, timing, which hand to support, and leg movements). This inspires us to pre-learn such a motion habit as motion prior, playing as supplement cues to refine PI-HMR's prediction. 

We apply a VQ-VAE as the motion prior learner. The selection is rooted in our assumption that the distribution of bed-bound movements is rather constrained. In that case, for a noised motion prediction, VQ-VAE could match it to the closest pattern, thereby re-generating habit-based results. The VQ-VAE is based on Transformer blocks and show similar architecture with~\cite{feng2024stratified}. During training,  we only auto-reconstruct the pose sequences~($\theta$ in SMPL). More details are provided in Supplementary Materials.

The VQ-VAE will act as the only motion prior and supervision in our TTO routine. For terminological convenience, given a VQ-VAE $\mathbb{M}$ and PI-HMR initial predictions~$\Theta^0 = \{\theta^0_1, ... \theta^0_T\}$, the $i_{th}$ iteration objectives follows:
\begin{equation} 
\small
    \mathcal{L}_{TTO}^i = \mathcal{L}_{m}(\Theta^i, \Theta^0) + \mathcal{L}_{m}(\Theta^i, \mathbb{M}(\Theta^i)) + \mathcal{L}_{sm}(\Theta^i)
\end{equation}
$\mathcal{L}_{m}$ is the SMPL and joint error term, and $\mathcal{L}_{sm}$ is the smooth loss. The result of $i_{th}$ iteration will be input into $\mathbb{M}$ and optimized in the ${i + 1}_{th}$ iteration. The TTO will help maintain a balance between initial PI-HMR outputs and the reconstruction by VQ-VAE, thus learning robust motion priors.



\section{Experiments}

We evaluate PI-HMR on the TIP dataset. Following~\cite{wu2024seeing},  we choose the second-to-last group of each subject as the val. set, the last group of each subject as the test set, and the remains as the training set. For evaluation, We use standard evaluation metrics including MPJPE~(without pelvis alignment), PA-MPJPE, MPVE for shape errors, and Acceleration errors~(ACC-ERR) to evaluate smoothness. The first three metrics are measured in millimeters~($mm$), and the rest are measured in $mm/s^2$.

We compare our model with previous SOTAs and vison-based classic structures, including: HMR~\cite{kanazawa2018end} and HMR-KD (HMR structure with and without cross-modal KD), BodyMap-WS~\cite{tandon2024bodymap}, TCMR~\cite{choi2021beyond}, MPS-NET~\cite{wei2022capturing}, and PI-Mesh~\cite{wu2024seeing}. All methods are re-trained on TIP with our re-generated SMPL p-GTs, and follow the same training setups with PI-HMR. We provide detailed implementation details of these approaches and PI-HMR in Sup. Mat.  

\subsection{Overall Results for PI-HMR} \label{sec:exp_overall_results}
We present quantitative evaluations in~\cref{tab: overall_results}. Our methods outperform all image or sequence-based methods, presenting about 17.01mm MPJPE decrease compared to PI-Mesh and also outperforms SOTA vision-based architecture HMR, TCMR with 15.6mm, 4.91mm MPJPE improvement, while maintaining comparable ACC-ERR compared with SOTA approaches. Moreover, our introduced cross-modal KD and TTO strategy further improve the robustness of PIHMR, bringing 2.33mm and 1.7mm MPJPE improvements compared with basic structure. In particular, the TTO strategy, as an unsupervised, entirely prior-based optimization strategy, demonstrates the effectiveness of learning and refinement based on user habits. We provide visual comparisons between CLIFF, PI-Mesh and PI-HMR in~\cref{fig: overall_vis}. 

\subsection{Ablations for PI-HMR}

In this section, we present various ablation studies to fully explore the best setup of PI-HMR. We select PI-HMR as shorthand to mean PI-HMR + KD, without the TTO routine, as the basic model for evaluation. All models are trained and tested with the same data as PI-HMR. 

\begin{table}[]
\footnotesize
\centering 
\begin{tabular}{cccc|cc}
\toprule[2pt]
GF & LF & SF-P & SF-K & MPJPE & PA-MPJPE\\
\midrule[1.4pt]
 \checkmark  &  \checkmark  &      &      &   57.84  &  43.18 \\
 \checkmark  &    &  \checkmark    &      &  59.26   &  45.27\\
 \checkmark &    &      &    \checkmark  &   58.31  &  43.92\\
 \checkmark  &    &    \checkmark  &  \checkmark    &  59.03  &  44.45  \\
 \cline{1-4}
 \checkmark  &  \checkmark  &   \checkmark   &      &   62.23  &  44.91 \\
 \checkmark  &  \checkmark  &      &   \checkmark   &   58.48  &  44.27\\
 \checkmark  &   \checkmark &    \checkmark  &  \checkmark    &  \textbf{57.13}  &  \textbf{42.98} \\
\bottomrule[2pt]
\end{tabular}
\caption{\textbf{Ablations for model structures}. GF, LF, SF-P, SF-K are the global features, local features, sampling features from high-pressure areas and joints, respectively.} \label{tab: res_comp_eff}
\end{table}

\textbf{Model Structures.} In~\cref{tab: res_comp_eff}, we summarize the results with different feature combinations in the MFF module. The method that integrates all branches surpasses other setups. Notably, we observe accuracy drops when sampling features are solely sampled from high-pressure areas, without joints. This could be attributed to the model's tendency to focus more on high pressure, neglecting the local distribution in boardline areas and low-pressure regions related with joints, thereby failing due to information ambiguity.

\textbf{Top-K Sampling.} We explore the rational selection K for the high-pressure masks in~\cref{tab: ablations for topk}. With an increase number of sampling points, the model's performance initially improves and then declines when K is 256. This implies that the model seeks a balance in multi-feature fusion: more sampling points entail more abundant contact and contour information and a broader field of perception, but bringing in redundancy and noises.

\textbf{Other Components in MFF.} We also conducted experiments to evaluate three essential modules including AttentionPooling for local features, learnable masks and spatial position embedding in MFF, as shown in~\cref{tab: ablations for PI-HMR}. Our results suggest that these components provide strong priors for supervision and significantly improve the prediction accuracy.

\begin{table}[]
    \small
    \centering
    \begin{tabular}{c|cc}
    \toprule[2pt]
    Sampling Method      & MPJPE  & PA-MPJPE \\
    \midrule[1.4pt]
    Top 8       &    58.62   &   44.42   \\
    Top 32      &    57.66   &   43.48   \\
    Top 128     &    \textbf{57.13}   &   \textbf{42.98}   \\
    Top 256     &   58.64    &  44.65    \\
    \bottomrule[2pt]
\end{tabular}
\caption{\textbf{Ablations for the K selection in TopK algorithm.}} \label{tab: ablations for topk}
\end{table}

\begin{table}[] \vspace{-0.08cm}
    \footnotesize
    \centering
    \begin{tabular}{l|cc}
    \toprule[2pt]
    Method      & MPJPE  & PA-MPJPE \\
    \midrule[1.5pt]
    w/o. Learnable Masks            &   60.95    &    46.27  \\
    w/o. Spatial Position Embedding           &   60.65    &   46.28   \\
    w/o. AttentionPooling            &    59.21   &  45.08    \\
    \hline
    All               &    \textbf{57.13}   &  \textbf{42.98}    \\
    \hline
    \bottomrule[2pt]
    \end{tabular}
    \caption{\textbf{Ablations for other components in MFF.}} \label{tab: ablations for PI-HMR}
    \vspace{-0.1cm}
\end{table}





\textbf{Ablations for KD.} We conduct experiments to evaluate cross-modal KD. \cref{tab: ablations_kd} shows that feature-based transfer plays a pivotal role in enhancing the performance, while CLIFF's results might, to some extent, misguide the learning of HMR, due to domain gaps~(CLIFF's encoder is pre-trained on ImageNet). When both supervisions coexist, HMR could learn the complete cognitive thought-chain of CLIFF, leading to refinement in predictions.

\begin{table}[] \vspace{-0.08cm}
    \footnotesize
    \centering
    \begin{tabular}{ccc|cc} 
    \toprule[2pt]
    GT         & Output-KD   & Feat.-KD       & MPJPE  & PA-MPJPE\\
    \midrule[1.4pt]
    \checkmark &            &            &  75.06     & 57.97 \\
    \checkmark & \checkmark &            &   77.86  & 59.41\\
    \checkmark &            & \checkmark &  67.34     &  52.16 \\
    \checkmark & \checkmark & \checkmark &  \textbf{66.3}    &  \textbf{52.41}\\
    \bottomrule[2pt]
    \end{tabular} 
    \caption{\textbf{Ablations for cross-modal KD}. GT, Output-KD, and Feat-KD represent supervision with GTs, CLIFF's outputs, and CLIFF's hidden feature maps, respectively.} \label{tab: ablations_kd} \vspace{-0.08cm}
\end{table} 

\subsection{Results for SMPLify-IB} \label{subsec:exp_re_SIB} 
\begin{table}[t]
    \footnotesize
    \centering
    \begin{tabular}{l||cc}
    \toprule[2pt]
        ~ & 2D MPJPE & Limb height \\ \midrule[1.4pt]
        CLIFF & 25.20 & - \\ \hline
        TIP & 14.02 & 142.84  \\ \hline
        SMPLify-IB & \textbf{9.65} & \textbf{66.68}  \\ 
    \bottomrule[2pt]
    \end{tabular}
    \caption{\textbf{Qualitative results for SMPLify-IB, compared with the p-GTs in TIP, and CLIFF's outputs}. We calculate the 2D projection errors~(in pixels), and the average height of limbs marked as stationary relative to the bed.} \label{tab:results_dataset}
\end{table}
\cref{tab:results_dataset} provides the evaluation of p-GTs generated by SMPLify-IB. Besides the 2D projection errors and acceleration metrics, we introduce the static limb height as an objective assessment of our refinement in implausible limb lifts. Given the prevalence of limbs placed on other body parts within TIP, this metric can only serve as a rough estimate under limited self-penetration premise. We provide visual results in the Sup. Mat. to present our enhancements.

\begin{table}[t] \vspace{-0.08cm}
    \footnotesize
    \centering
    \begin{tabular}{l||cccc}
    \toprule[2pt]
        ~ & recall & precision & accuracy & time  \\ \midrule[1.4pt]
        SMPLify-XMC & 100\% & 100\% & 100\% & 22.62s  \\ \hline
        Ours & 70.93\% & 80.64\% & 98.32\% & 0.42s  \\ \hline
        Ours (ds 1/3) & 65.66\% & 73.59\% & 98.03\% & 0.036s  \\ 
    \bottomrule[2pt]
    \end{tabular}
    \caption{\textbf{Comparisons between our penetration detection algorithm with SMPLify-XMC.} Time means time consumption in an iteration when deploying detection algorithms in our optimization. 'ds 1/3' means downsample SMPL vertices to their 1/3 scales. } \label{tab: res_detection}
\end{table}

We use SMPLify-XMC's detection results as the GTs and conduct comparison experiments to evaluate our light-weight self-penetration detection algorithm in~\cref{tab: res_detection}. The experiment run on the first group of the TIP dataset. For each batch with 128 images, we integrate both detection algorithms in our optimization routine, record the runtime for each iteration~(1000 iterations for a batch) and calculate the accuracy, precision, and recall of the detection. Compared with SMPLify-XMC, our detection module achieves 53.9 times faster while maintaining a detection accuracy of 98.32\%. We also implement a more lightweight version by downsampling the SMPL vertices into their 1/3 scale. The downsampled version further yields a more than tenfold increase in speed, accompanied by limited precision decrease. 


%


\section{Conclusion}

In this work, we present a general framework for in-bed human shape estimation with pressure images, bridging from pseudo-label generation to algorithm design. For label generation, we present SMPLify-IB, a low-cost monocular optimization approach to generate SMPL p-GTs for in-bed scenes. By introducing gravity constraints and a lightweight but efficient self-penetration detection module, we regenerate higher-quality SMPL labels for a public dataset TIP. For model design, we introduce PI-HMR, a pressure-based HPS network to predict in-bed motions from pressure sequences. By fusing pressure distribution and spatial priors, accompanied with KD and TTO exploration, PI-HMR outperforms previous methods. Results verify the feasibility of enhancing model's performance by exploiting pressure's nature.

\setcounter{section}{0}
\renewcommand{\thesection}{\Alph{section}}
\clearpage
\setcounter{page}{1}
\maketitlesupplementary

\begin{figure*}[h]
  \centering
  \includegraphics[width=\linewidth]{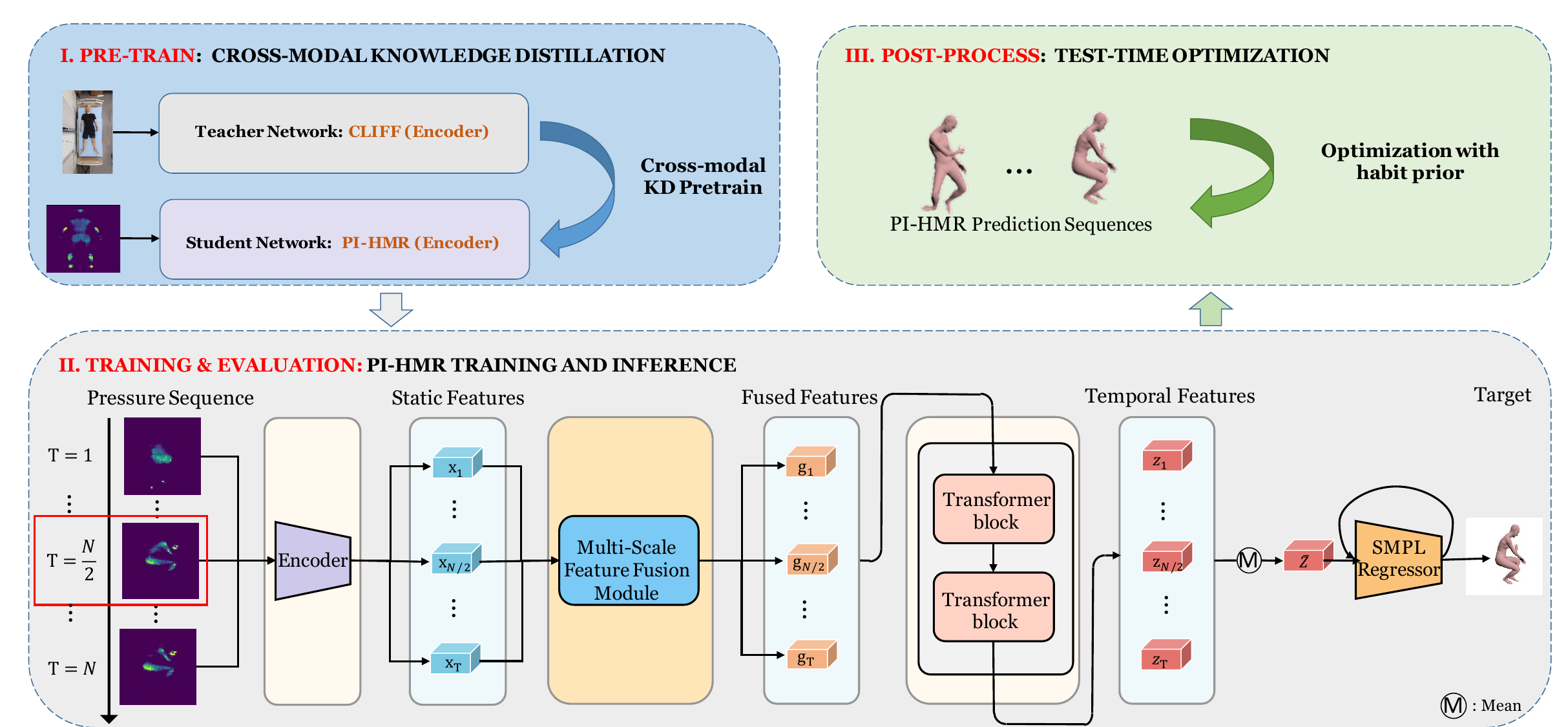}
  \caption{Our data flow includes three stages: (1) pre-train: knowledge distillation-based cross-modal pre-training; (2) train \& evaluation: train the PI-HMR network with pressure sequences; (3) post-process: improve the estimates with the Test-Time Optimization strategy.} 
  \label{fig: sup_overall_pipeline}
\end{figure*}

\section{Introduction}
In this material, we provide additional details regarding the network and implementation of our methods, as well as compared SOTAs. We further present more qualitative results to show the performance of PI-HMR and our re-generated p-GTs for TIP~\cite{wu2024seeing} and to explore their failure scenarios. The details include:
\begin{itemize}
    \item Implementation details for SMPLify-IB, PI-HMR, cross-modal knowledge distillation, VQ-VAE, test-time optimization, and SOTA methods compared to PI-HMR.
    \item More quantitative and qualitative results about SMPLify-IB, PI-HMR, and failure cases.
    \item Limitations and future works.
\end{itemize}

The overall pipeline of our pressure-to-motion flow is shown in~\cref{fig: sup_overall_pipeline}, and detailed architecture and implementation details will be elaborated below.

\section{Preliminary}

\textbf{Body Model.} The SMPL~\cite{loper2023smpl} model provides a differentiable function $V = \mathcal{M}(\theta, \beta, t)$ that outputs a posed 3D mesh with $N=6890$ vertices. The pose parameter $\theta \in \mathbb{R}^{24 \times 3}$ includes a $\mathbb{R}^3$ global body rotation and the relative rotation of 23 joints with respect to their parents. The shape parameter $\beta \in \mathbb{R}^{10}$ represents the physique of the body shape. And $t \in \mathbb{R}^3$ means the root translation w.r.t the world coordinate.

\section{Network and Implementation Details}

\subsection{Implementation details for SMPLify-IB}

\label{sec:appendix_section}
\subsubsection{The first stage}
In the first stage of our optimization algorithm, we jointly optimize body shape $\beta$, pose parameters $\theta$, and translation $t$ using a sliding-window~(set as 128) approach, with overlap~(set as 64) between adjacent windows. We minimize the following objective function:
\begin{align}
L_{s1}(\theta, \beta, t)&=\lambda_{J}\mathcal{L}_{J}+
\lambda_{p}\mathcal{L}_{p}+\lambda_{sm}\mathcal{L}_{sm}+\lambda_{cons}\mathcal{L}_{cons}\nonumber\\
&\quad+\lambda_{bc}\mathcal{L}_{bc}+\lambda_{g}\mathcal{L}_{g}+\lambda_{sc}\mathcal{L}_{\sc}
\end{align}

1. \textbf{Reprojection constraint term $\mathcal{L}_{J}$:} This term penalizes the weighted robust distance between the projections of the estimated 3D joints and the annotated 2D joint ground truths. Instead of the widely used weak-perspective projection in~\cite{bogo2016keep} with presumed focal length, we apply the perspective projection with calibrated focal length and camera-bed distance provided by TIP.

2. \textbf{Prior constraint term $\mathcal{L}_{p}$:} This term impedes the unrealistic poses while allowing possible ones. $\mathcal{L}_{pose}$, $\mathcal{L}_{shape}$ penalizes the out-of-distribution estimated postures and shapes, which is similar to terms in SMPLify, and $\mathcal{L}_{torso}$ ensures correct in-bed torso poses, where the height of hips should be less than shoulders and the height of waist is below the mean height of shoulders and hips.
\begin{align}
\mathcal{L}_{p}&= \mathcal{L}_{pos} + \mathcal{L}_{sha} + \mathcal{L}_{tor} \\
\mathcal{L}_{pos} &= \sum^{T}_{i}(\lambda^{pos}_{1} (\mathcal{G}(\theta(i)) + \sum_j \lambda^{pos}_{2, j} \cdot e^{\gamma_j \cdot \theta(i)_j})) \nonumber \\
\mathcal{L}_{sha} &= \lambda^{sha}_{} \sum^{T}_{i}||\beta(i)||^2 \nonumber \\
\mathcal{L}_{tor} &= \sum^T_i (\lambda^{tor}_{1} \cdot e^{\omega_{hip}d_{hip}(i)}+ \lambda^{tor}_{2} \cdot e^{\omega_{wai}d_{wai}(i)}) \nonumber \\
d_{hip}(i) &= z_{hip}(i) - z_{sho}(i) \nonumber \\
d_{wai}(i) &= z_{wai}(i) - mean(z_{hip}(i), z_{sho}(i)) \nonumber
\end{align}
where $\mathcal{G}$ is the Gaussian Mixture Model pre-trained in SMPLify, and the second term in $\mathcal{L}_{pos}$ penalizes impossible bending of limbs, neck and torso, such as shoulder twist.  $z_{hip}$, $z_{sho}$, $z_{wai}$ are the height of hip joints, shoulder joints, and waist joint, and $\omega_{hip}$, $\omega_{wai}$ are both set to 100.

3. \textbf{Smooth constraint term $\mathcal{L}_{sm}$:} This term reduces the jitters by minimizing the 3D joints velocity, acceleration and SMPL parameter differences.
\begin{align}
\mathcal{L}_{smo}&=\mathcal{L}_{par}+ \mathcal{L}_{vel}+\mathcal{L}_{acc}  \\
\mathcal{L}_{par}&=\sum^{T-1}_{i=1} (\lambda^{par}_{1} ||\beta (i+1)-\beta (i)||^2 \nonumber \\
&\quad+ \lambda^{par}_{2} ||\theta (i+1)-\theta (i)||^2+\lambda^{par}_{2} ||t (i+1)-t(i)||^2) \nonumber \\
\mathcal{L}_{vel}&=\sum^{T-1}_{i=1}(\lambda^{vel}_{1} ||J(i+1)_{3D}-J(i)_{3D}||^2  \nonumber\\
&\quad+ \lambda^{vel}_{2} ||V(i+1)-V(i)||^2)  \nonumber \\
\mathcal{L}_{acc}&=\sum^{T-1}_{i=2} ||2J(i)_{3D}-J(i-1)_{3D}-J(i+1)_{3D}||^2 \nonumber
\end{align}
where $V(i)$ and $J(i)$ are the coordinates of SMPL vertex set $V$ and 3D joints $J$ in the frame $i$. 

4. \textbf{Consistency constraint term $\mathcal{L}_{cons}$:} This term enhances the consistency between the overlapping parts of the current window and the previously optimized window.
\begin{align}
\mathcal{L}_{cons}&=\sum_{\substack{i\in overlap \\ frames}} (\lambda^{cons}_{1}||\theta(i, b_{1})-\theta(i, b_{2})||^2  \nonumber\\
&\quad+\lambda^{cons}_{2} ||t(i, b_{1})-t(i, b_{2})||^2  \nonumber \\
&\quad+\lambda^{cons}_{3} ||V(i, b_{1}) - V(i, b_{2})||^2 \nonumber \\
&\quad+\lambda^{cons}_{4} ||J(i, b_{1})_{3D} - J(i, b_{2})_{3D}||^2) 
\end{align}
where $t(i, b)$, $\theta(i, b)$ is the translation parameters and pose parameters in frame $i$ of window $b$, and  $V(i, b)$, $J(i, b)_{3D}$ is the coordinates of vertex set $V$, 3D joints $J$ in frame $i$ of window $b$. $b_1$, $b_2$ means the previous window and the present window, respectively.

5. \textbf{Bed contact constraint term $\mathcal{L}_{bc}$:} This term improves the plausibility of human-scene contact. We consider vertices that are close to the bed to be in contact with bed and encourage those vertices to contact with the bed plane while penalizing human-bed penetration.
\begin{align}
\mathcal{L}_{bc} &= \sum^{T}_{i} (\lambda^{in\_bed}_{}  \sum_{0<z(i)_{v}<thre_{bed}} \text{tanh}^2(\omega_{in\_bed} z(i)_v) \nonumber\\ 
&\quad+ \lambda^{out\_bed}_{} \sum_{z(i)_{v}<0} \text{tanh}^2(-\omega_{out\_bed} z(i)_{v})) 
\end{align}
where $z(i)_v$ is the signed distance to the bed plane of vertex $v$ in frame $i$, and $thre_{bed}$ is the contact threshold and set to 0.02m.

6. \textbf{Gravity constraint term $\mathcal{L}_{g}$:} This term penalizes abnormal limb-lifting and reduces depth ambiguity.
\begin{align}
\mathcal{L}_{g}&=\sum^{T}_{i}\sum_{\substack{j\in G_{J}\\z(i)_j>0}}\mathbb{I}(vel(i)_{j}< thre_{vel})e^{\omega(i)_{j}(z(i)_j)} 
\end{align}

where $thre_{vel}$ is set to $\sqrt{110}$, and $vel(i)_j$ denotes the velocity of joint $j$ in frame $i$, which is calculated from 2D annotations. $\omega(i)_j$ is a dynamic weight depends on the state of annotated 2D joint ground truths. Specifically, in addition to the velocity-based criterion, we have more complicated settings for potential corner cases. For example, when a person is seated on the bed, supporting the bed surface with both hands, the shoulders will be incorrectly judged as implausible lifts by sole velocity-based criterion~(This scenario is rarely encountered in TIP, yet it still exists). In that case, we alleviate the impact of gravity constraints on this scenario by dynamically adjusting $\omega(i)_j$. In practice, when the 2D projection lengths of limbs are less than 60\% of the projection lengths in the rest pose, according to geometry, we consider the corresponding limb to be normally lifted even if the corresponding joint speed is below $thre_{v}$, and thus $\omega(i)_j$ takes a smaller value. Besides, $\omega(i)_j$ takes a smaller value for hand joints whose 2D projections are inside the torso to avoid severe hand-torso intersection.

7. \textbf{Self-contact constraint term $\mathcal{L}_{sc}$:} This term is proposed to obtain plausible self-contact and abbreviate self-penetration. In the first stage, we only deal with the intersection between the hand and the torso. The self-contact between other body parts is optimized in the second stage.
\begin{align}
\mathcal{L}_{sc}&=\lambda^{p\_con}_{} \mathcal{L}_{p\_con}+ \lambda^{p\_isect}_{} \mathcal{L}_{p\_isect}+ \lambda^{pull}_{} \mathcal{L}_{pull}  \nonumber \\
&\quad+ \lambda^{push}_{} \mathcal{L}_{push}  \\
\mathcal{L}_{p\_con}&=\sum_{\substack{0<sdf_v<thre_{dist}}} \text{tanh}^2(\omega_{p\_con}sdf_v) \nonumber \\
\mathcal{L}_{p\_isect}&=\sum_{sdf_v<0} \text{tanh}^2(\omega_{p\_isect}|sdf_v|) \nonumber
\end{align}
where $sdf_v$ is the value of the signed distance field(SDF) at vertex $v$, which is calculated by our self-penetration detection algorithm. The details of $\mathcal{L}_{pull}$, $\mathcal{L}_{push}$ are given in the main body of the manuscript.

\subsubsection{The second stage}
We treat the results of the first stage as initialization for the second stage. Specifically, we use the mean $\beta$ of each subject and fix the shape parameters in the second stage. We optimize $\theta$ and $t$ to obtain more plausible human meshes. The objective function $L_{s2}$ is as follows:
\begin{align}
L_{s2}(\theta, \beta, t)&=\lambda_{J}\mathcal{L}_{J}+
\lambda_{p}\mathcal{L}_{p}+\lambda_{sm}\mathcal{L}_{sm}+\lambda_{cons}\mathcal{L}_{cons}\nonumber\\
&\quad+\lambda_{bc}\mathcal{L}_{bc}+\lambda_{g}\mathcal{L}_{g}+\lambda_{sc}\mathcal{L}_{sc}
\end{align}
$\mathcal{L}_J$, $\mathcal{L}_p$, $\mathcal{L}_{sm}$, $\mathcal{L}_{cons}$, $\mathcal{L}_{g}$, $\mathcal{L}_{bc}$ are the same as the first stage, while $L_{sc}$ penalizes self-intersection in all body segments rather than only hands and torso.

\subsubsection{Implementation details}
We use Adam as the optimizer with a learning rate of 0.01, and each stage involves 500 iterations. The length of the sliding window is 128, with 50\% overlapping to prevent abrupt changes between windows. The joints and virtual joints we use for the segmentation of SMPL mesh is listed in~\cref{tab:opt_SEG}.
\setlength{\aboverulesep}{0pt}
\setlength{\belowrulesep}{0pt}
\begin{table}[]
\centering
\begin{tabular}{m{7em}|m{15em}}
\toprule[2pt]
body parts & joints \& virtual joints \\
\midrule[1.2pt]
head & left ear, right ear, nose \\
\hline
torso-upper arm & left shoulder, right shoulder, spine2 \\
\hline
\par left arm & left elbow, left hand, \par mid of left elbow and hand, \par $\frac{2}{5}$ point from left elbow to shoulder \\
\hline
\par right arm & right elbow, right hand, \par mid of right elbow and hand, \par $\frac{2}{5}$ point from right elbow to shoulder \\
\hline
torso-thigh & left hip, right hip \\
\hline
\par left thigh & left knee, left ankle, \par mid of left knee and ankle, \par $\frac{2}{5}$ point from left ankle to hip \\
\hline
\par right thigh & right knee, right ankle, \par mid of right knee and ankle, \par$\frac{2}{5}$ point from right ankle to hip \\
\bottomrule[2pt]
\end{tabular}
\caption{\textbf{Positions of our selected segment centers}.}
\label{tab:opt_SEG}
\end{table}

\subsection{Implementation details for PI-HMR}

Before the aforesaid modules in the main body of our manuscript, PI-HMR also contains three different Transformer blocks for AttentionPooling, cross-attention for sampling features in MFF, and temporal consistency extraction. We will provide detailed designs of these Transformer layers.  (1) For AttentionPooling, we use the same structure in CLIP~\cite{radford2021learning}. (2) For the cross-attention module in MFF, we apply a one-layer Transformer block as the attention module with one attention head and  Dropout set as 0. (3) For the temporal encoder, we apply a two-layer Transformer block to extract the temporal consistency from the fusion feature sequence. In detail, each transformer layer contains a multi-head attention module with $N=8$ heads. These learned features are then fed into the feed-forward network with 512 hidden neurons. Dropout~($p=0.1$) and DropPath~($p_d=0.2$) are applied to avoid overfitting.

The loss of PI-HMR is defined as:
\begin{equation} \label{eq: sup_overall_func}
\small
   \mathcal{L}_{pi} = \lambda_{\text{SMPL}} \mathcal{L}_{\text{SMPL}} + \lambda_{3D} \mathcal{L}_{3D} + \lambda_{2D} \mathcal{L}_{2D} 
\end{equation}
where $\mathcal{L}_{\text{SMPL}}$, $\mathcal{L}_{3D}$, $\mathcal{L}_{2D}$ are calculated as:

\begin{equation}
\mathcal{L}_{\text{SMPL}} = \omega_{s}^{\text{SMPL}} ||\beta - \hat{\beta}||^2 + \omega_{p}^{\text{SMPL}} ||\theta - \hat{\theta}||^2  + \omega_{t}^{\text{SMPL}} ||t - \hat{t}||^2 \nonumber 
\end{equation}

\begin{equation}
\mathcal{L}_{3D} = || J_{3D} - \hat{J}_{3D} ||^2 \nonumber 
\end{equation}

\begin{equation}
\mathcal{L}_{2D} = || J_{2D} - \hat{J}_{2D} ||^2 \nonumber 
\end{equation}
where $\hat{x}$ represents the ground truth for the corresponding estimated variable $x$, and $\lambda$ and $\omega$ are hyper-parameters. We set $\lambda_{\text{SMPL}}=1$, $\lambda_{3D}=300$, $\lambda_{2D}=0.5$, $\omega_{t}^{\text{SMPL}}=\omega_{p}^{\text{SMPL}}=60$, and $\omega_{s}^{\text{SMPL}}=1$ for PI-HMR's training.

Before training, we first pad pressure images to $64 \times 64$ and set $T=15$ as the sequence length. No data augmentation strategy is applied during training. During the training process, we train PI-HMR for 100 epochs with a batchsize of 16, using the AdamW optimizer with a learning rate of 3e-4 and weight decay of 5e-3. We adopt a warm-up strategy in the initial 5 epochs and schedule periodically in a cosine-like function as~\cite{wu2024seeing}. The weight decay is set to 5e-3 to abbreviate overfitting. All implementation codes are implemented in the PyTorch 2.0.1 framework and run on an RTX4090 GPU. 

\subsection{Implementation details for cross-modal KD}\label{sec:sup_imp_kd}
\begin{figure*}[htbp]
  \centering
  \includegraphics[width=\linewidth]{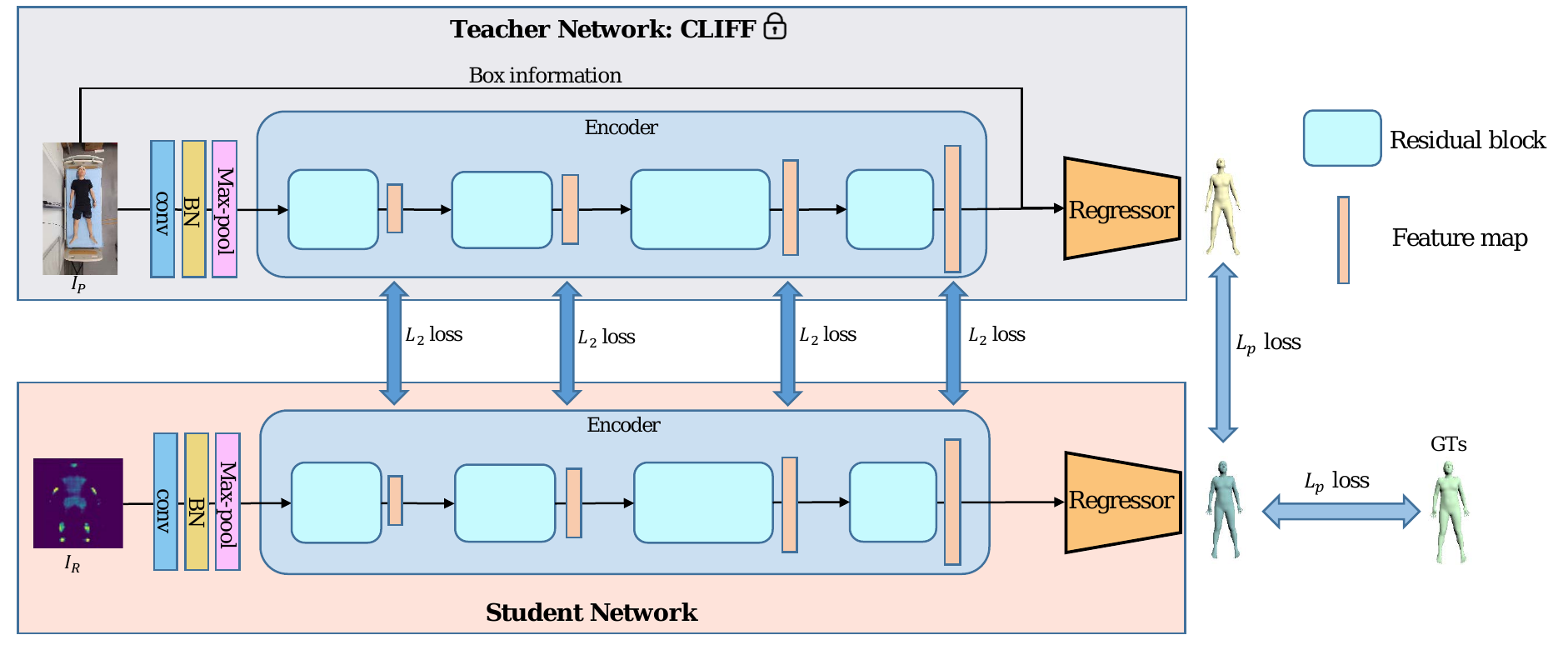}
  \caption{\textbf{An overview of our KD-based network.}} 
  \label{fig: sup_kd_structure}
\end{figure*}
We conduct a HMR-based network~(with a ResNet50 as encoder and an IEF~\cite{kanazawa2018end} SMPL regressor) to pre-train the ResNet50 encoder with SOTA vision-based method Cliff~\cite{li2022cliff}. The detailed structure is presented in~\cref{fig: sup_kd_structure} where we concurrently introduce label supervision, as well as distillation from Cliff's latent feature maps and prediction outcomes, to realize cross-modal knowledge transfer.

To train the KD-based network, like PI-HMR, we first pad pressure images to $64 \times 64$. No data augmentation strategy is applied during training. The training process is performed for 100 epochs with an AdamW optimizer in a minibatch of $256$ on the same training and validation dataset of PI-HMR. We adopt a warm-up strategy in the initial 5 epochs and schedule periodically in a cosine-like function. The weight decay is set to 5e-3 to abbreviate overfitting. All implementation codes are implemented in the PyTorch 2.0.1 framework and run on an NVIDIA. RTX4090 GPU. 

\subsection{Implementation details for VQ-VAE}
\begin{figure}[htbp]
  \centering
  \includegraphics[width=\linewidth]{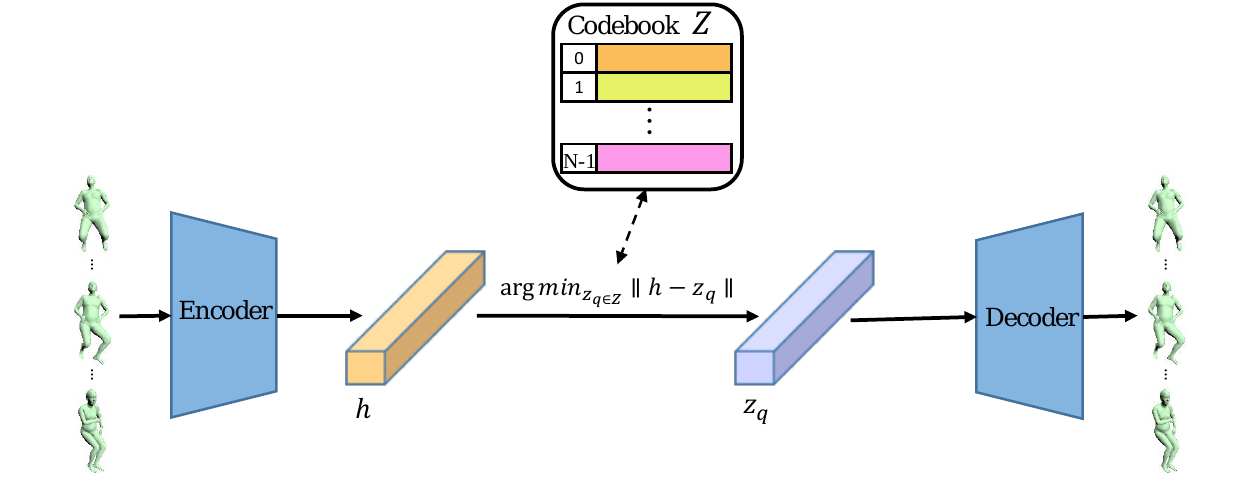}
  \caption{\textbf{An overview of our VQ-VAE network.}} 
  \label{fig: sup_kd_structure}
\end{figure}
The VQ-VAE follows the architecture in~\cite{feng2024stratified}, which incorporates two 4-layer Transformer blocks as the encoder and decoder, respectively, and a $\mathbb{R}^{512 \times 384}$ codebook with 512 entries and $\mathbb{R}^{384}$ for the discrete latent of each entry. Each Transformer layer consists of a 4-head self-attention module and a feedforward layer with 256 hidden units.

To train the VQ-VAE network, we only input the pose parameter sequence $\Theta=\{\theta_1, ..., \theta_T\}$, without the translation and shape parameters, to push the model learning the motion continuity of the turn-over process. The pose sequences will firstly be encoded to motion features $H$ in the Transformer encoder, quantized into discrete latent sequence $Z$ by finding its closest element in the codebook, and reconstruct the input motion sequence in the follow-up Transformer decoder. We follow the loss setting in~\cite{feng2024stratified} and minimize the following loss function in~\cref{eq: sup_loss_vqvae}.

\begin{align} \label{eq: sup_loss_vqvae}
\mathcal{L}_{vq} = \lambda^{vq}_{\theta}&\text{Smooth}_{L1}(\Theta, \hat{\Theta}) \\
&\quad+ \lambda^{vq}_{J}\text{Smooth}_{L1}(J_{3D}(\Theta), J_{3D}(\hat{\Theta})) \nonumber\\
&\quad+ \lambda^{vq}_{d}(||sg[Z] - H||_2 + \omega^{vq}_{b} ||Z - sg[H]||_2) \nonumber
\end{align}
where $\hat{x}$ represents the ground truth for the corresponding estimated variable $x$, $\mathcal{J}(\Theta)$ means 3D joint locations of given SMPL pose parameter sequences $\Theta$~($\beta$ and $t$ are default all-0 tensors), $sg$ denotes the stop gradient operator, and $\lambda$ and $\omega$ are hyper-parameters. We set $\lambda^{vq}_{\theta}=1$, $\lambda^{vq}_{J}=5$, $\lambda^{vq}_{d}=0.25$, and $\omega^{vq}_{b}=0.5$. 

The VQ-VAE is trained with a batchsize of 64 and a sequence length of 64 frames for 100 epochs on the same training and validation dataset of PI-HMR. Adam optimizer is adapted for training, with a fixed learning rate of 1e-4, and [0.9, 0.999] for $\beta$ of the optimizer. All implementation codes are implemented in the PyTorch 2.0.1 framework and run on an NVIDIA. RTX4090 GPU. 

\subsection{Implementation details for Test-Time Optimization}

We use the VQ-VAE to act as the only motion prior and supervision in our TTO routine. For terminological convenience, given a VQ-VAE $\mathbb{M}$ and PI-HMR initial predictions~$\Theta^0 = \{\theta^0_1, ... \theta^0_T\}$. For the $i_{th}$ iteration, we calculate the loss by~\cref{eq: sup_tto} and update the $\Theta$ by stochastic gradient descent. The result of $i_{th}$ iteration will be input into $\mathbb{M}$ and optimized in the ${i + 1}_{th}$ iteration:

\begin{equation} 
    \mathcal{L}_{TTO}^i = \alpha \mathcal{L}_{m}(\Theta^i, \Theta^0) + (1 - \alpha)\mathcal{L}_{m}(\Theta^i, \mathbb{M}(\Theta^i)) + \mathcal{L}_{sm}(\Theta^i)
\label{eq: sup_tto}
\end{equation}
where each term is calculated as:

\begin{align} 
 \mathcal{L}_{m}(\Theta_1, \Theta_2) =  \lambda_{\text{smpl}}^{TTO}  ||\Theta_1, \Theta_2||^2 \\
&+ \lambda_{3D}^{TTO} ||\mathcal{J}(\Theta_1) - \mathcal{J}(\Theta_2)||^2 \nonumber
\end{align}

\begin{equation}
\begin{split}
    \mathcal{L}_{sm}(\Theta) = \lambda_{sm}^{TTO} \frac{1}{T-1} \sum_{t=2}^{T} (|\Theta(t) - \Theta(t-1)| \\
    + |\mathcal{J}(\Theta(t)) - \mathcal{J}(\Theta(t-1)| \nonumber
\end{split}
\end{equation}
where $\mathcal{J}(\Theta)$ means 3D joint locations of given SMPL pose parameters $\Theta$~($\beta$ and $t$ are the initial predictions and won't be updated during the optimization), $\alpha$ is a balance weight to balance initial PI-HMR predictions and reconstructions of VQ-VAE, and $\lambda$s are hyperparameters. We set $\alpha=0.5$, $\lambda_{\text{smpl}}^{TTO}=0.5$, $\lambda_{3D}^{TTO}=0.1$, and $\lambda_{sm}^{TTO}=0.1$ for the test-time optimization.

During the optimization, we freeze the shape parameters $\beta$ and translation parameters $t$ of the initial PI-HMR's outputs, and only optimize pose parameters $\theta$. We employ a sliding window of size 64 to capture the initial PI-HMR predictions and  update them in 30 iterations with a learning rate of 0.01 and Adam as the optimizer. All optimization codes are implemented in the PyTorch 1.11.0 framework and run on an NVIDIA. RTX3090 GPU.

\subsection{Implementation details for SOTA methods}

In this section, we will provide implementation details of compared SOTA networks.

\textbf{HMR~\cite{kanazawa2018end} and HMR + KD}: The implementation details of HMR series are introduced in~\cref{sec:sup_imp_kd}. The distinction between the two lies in whether knowledge distillation supervision is employed during the training process.

\textbf{TCMR~\cite{choi2021beyond} and MPS-NET~\cite{wei2022capturing}}: We choose TCMR and MPS-NET as the compared vision-based architecture because they follow the same paradigm of VIBE~\cite{kocabas2020vibe}, which incorporates a static encoder for texture feature extraction, a temporal encoder for temporal consistency digestion, and a regressor for final SMPL predictions. We use the same architecture and loss weights of the default setting, except converting the initial ResNet50 input to a single channel and adjusting the first convolution layer's kernel size to $5\times 5$ to fit the single-channel pressure images.

\textbf{PI-Mesh~\cite{wu2024seeing}}: PI-Mesh is the first-of-its-kind temporal network to predict in-bed human motions from pressure image sequences. We follow the codes and implementation details provided in~\cite{wu2024seeing} with a ResNet50 as the static encoder and a two-layer Transformer block as the temporal encoder.  

\textbf{BodyMAP-WS}: BodyMap~\cite{tandon2024bodymap} is a SOTA dual-modal method to predict in-bed human meshes and 3D contact pressure maps from both pressure images and depth images. We realize a substitute version provided in their paper, named BodyMap-WS, because we don't have 3D pressure map labels. It is worth mentioning that we notice the TIP dataset fails to converge on the algorithm provided in their GitHub repository. So we remove part of the codes including rotation data augmentation and post-processing of the network outputs~(Line 139-150 and Line 231-242 in the \textit{PMM/MeshEstimator.py} of the GitHub repository) to ensure convergence. 

All methods are trained on the same training-validation dataset of PI-HMR. For TCMR, MPS-NET, and PI-Mesh, we adopt the same training routine as PI-HMR. To be specific, we first pad pressure images to $64 \times 64$ and set $T=15$ as the sequence length. No data augmentation strategy is applied during training. During the training process, we train these approaches for 100 epochs with a batchsize of 16, using the AdamW optimizer with the learning rate of 3e-4 and weight decay of 5e-3~(we firstly conduct a simple grid-search for the best learning rate selection on these methods), and adopt a warm-up strategy in the initial 5 epochs and scheduled periodically in a cosine-like function. For BodyMap-WS, we follow the training routine provided in~\cite{tandon2024bodymap}, resize the pressure images to $224\times224$, and apply RandomAffine, RandomCutOut, and PixelDropout as data augmentation strategies. The training process is performed for 100 epochs with an Adam optimizer in a minibatch of $32$, a learning rate of 1e-4 and weight decay of 5e-4. All codes are implemented in the PyTorch 2.0.1 framework and run on an NVIDIA. RTX4090 GPU.

\section{More ablations}

\subsection{Discussion on TopK sampling} 
\begin{figure}[t]
  \centering
  \includegraphics[width=0.8\linewidth]{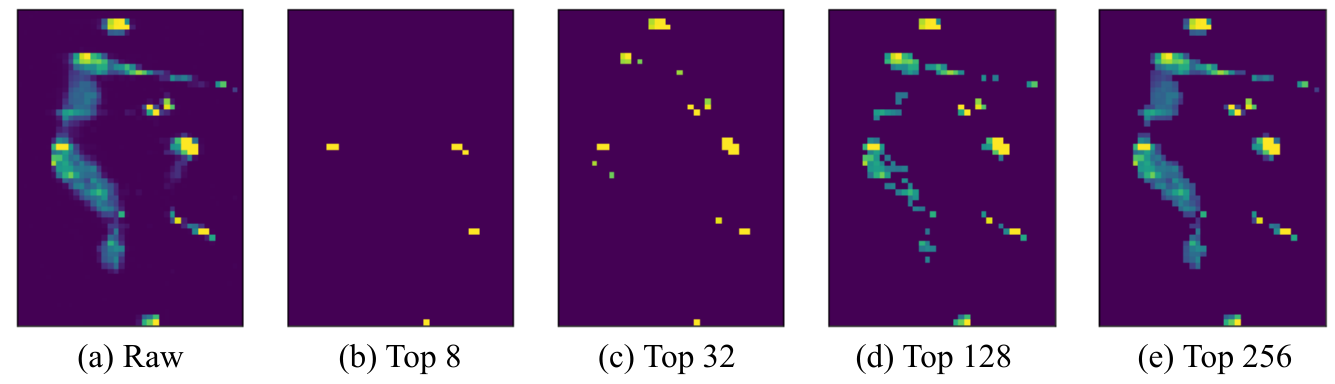} 
  \caption{\textbf{Visualization of TopK Sampling.}} 
  \label{fig: topk_sampling}
\end{figure}

The sampling functions as a low-value filter, freeing the model's attention from redundant, noisy backgrounds and focusing more on high-value regions. We provide a visualization in~\cref{fig: topk_sampling}, where, with 128 points, the pressure image can retain the human's outline while highlighting the core contact areas.

\subsection{Comparisons with single-input models} For vision methods, single-image models usually exhibit lower MPJPE compared to temporal models~(\eg CLIFF vs PMCE). However, for pressure data, temporal models show superiority, likely due to their ability to leverage temporal context, mitigating information ambiguity. This implies the strength of temporal models in pressure data processing compared to single ones. For fair comparisons, we implemented a single-input-based PI-HMR, achieving a 62.01mm MPJPE~(71.48mm for BodyMAP-WS), showing the efficacy of our architecture framework.

\subsection{Results on the original TIP dataset}
The results are shown in~\cref{tab: results_on_o_tip}, which demonstrate a comparable magnitude of MPJPE reduction, proving the efficacy of PI-HMR.

\begin{table}[t] 
\small
\centering
\begin{tabular}{l|c|c|c}
\hline
Method      & TCMR  & PI-Mesh & PI-HMR\\
\hline
MPJPE/ACC-ERR & 67.9/14.6 & 79.2/18.2 & \textbf{68.38/5.24}\\
\hline
\end{tabular} 
\caption{\textbf{Quantitative results on the original TIP dataset.}} \label{tab: results_on_o_tip} 
\end{table}

\subsection{Ablations of TTO.} 
We conducted ablations involving the selection of the balance weight $\alpha$ in~\cref{tab: ablations_alpha} and the number of iterations in~\cref{tab: ablations_iters}. We also explored integrating the pre-trained VQ-VAE into PI-Mesh during training~(as it regresses the sequence rather than the mediate frame, making it suitable for VQ-VAE) and calculating the reconstruction loss. However, MPJPE drops limitedly (0.06mm). We will explore more potential methods~(\eg SPIN-like) in the future work.

\begin{table}[] 
\centering
\begin{tabular}{l|ccccc}
\hline
$\alpha$      & 0.1  & 0.3 & 0.5 & 0.7 & 0.9\\
\hline
MPJPE      &   56.94  &   55.93   &   55.50   & \textbf{ 55.43}   &   55.67  \\
\hline
\end{tabular}
\caption{\textbf{Ablations on balance weight $\alpha$.}} 
\label{tab: ablations_alpha}
\end{table}

\begin{table}[] 
\centering
\begin{tabular}{l|ccccc}
\hline
iters      & 10  & 30 & 50 & 70 & 90\\
\hline
MPJPE      &  56.14   &  55.50    &   55.25   &  55.15   &   \textbf{55.10 } \\
\hline
\end{tabular}
\caption{\textbf{Ablations on the number of iterations.}} 
\label{tab: ablations_iters}
\end{table}

\subsection{Generalization of SMPLify-IB on the SLP dataset.}  
\begin{figure}[t]
  \centering
  \includegraphics[width=\linewidth]{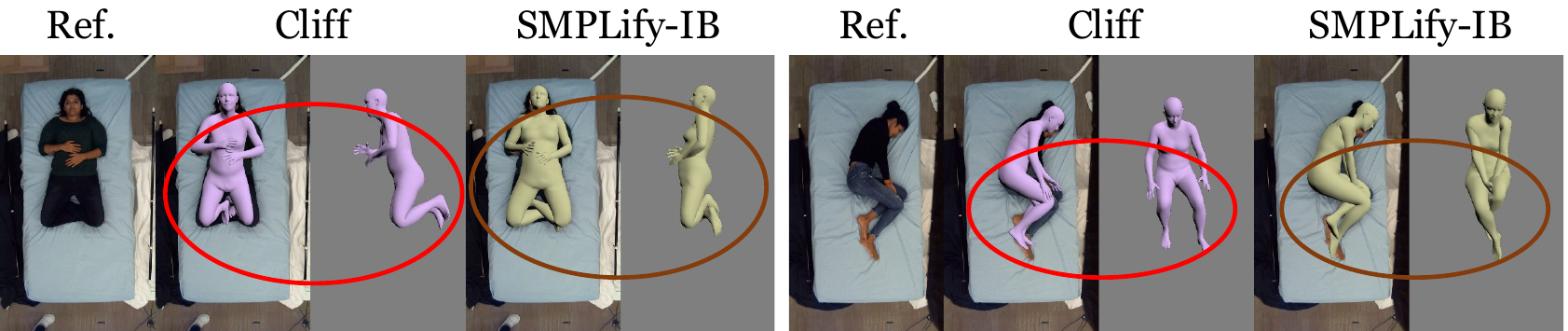}
  \caption{\textbf{Visualizations of SMPLify-IB on SLP.}} 
  \label{fig: ib_slp}
\end{figure}

We implemented SMPLify-IB on the SLP dataset. Results show the 2D MPJPE drops from 37.6 to 6.9 pixels compared to Cliff's outputs. \cref{fig: ib_slp} shows our pros in alleviating depth ambiguity. Meanwhile, we observed limb distortions in the optimization results, which may stem from erroneous initial estimations (CLIFF exhibits notable domain adaptation issues in an in-bed scene). In the absence of temporal context, these mis-predictions could exacerbate the likelihood of unreasonable limb angles, underscoring the significance of temporal information in in-bed human shape annotations.

\section{Visualization results}

In this section, we present additional visualization results to verify the efficiency of our general framework for the in-bed HPS task.
\subsection{Visualizations for Time Consumption of self-penetration algorithms}
\begin{figure}[htbp]
  \centering
  \includegraphics[width=\linewidth]{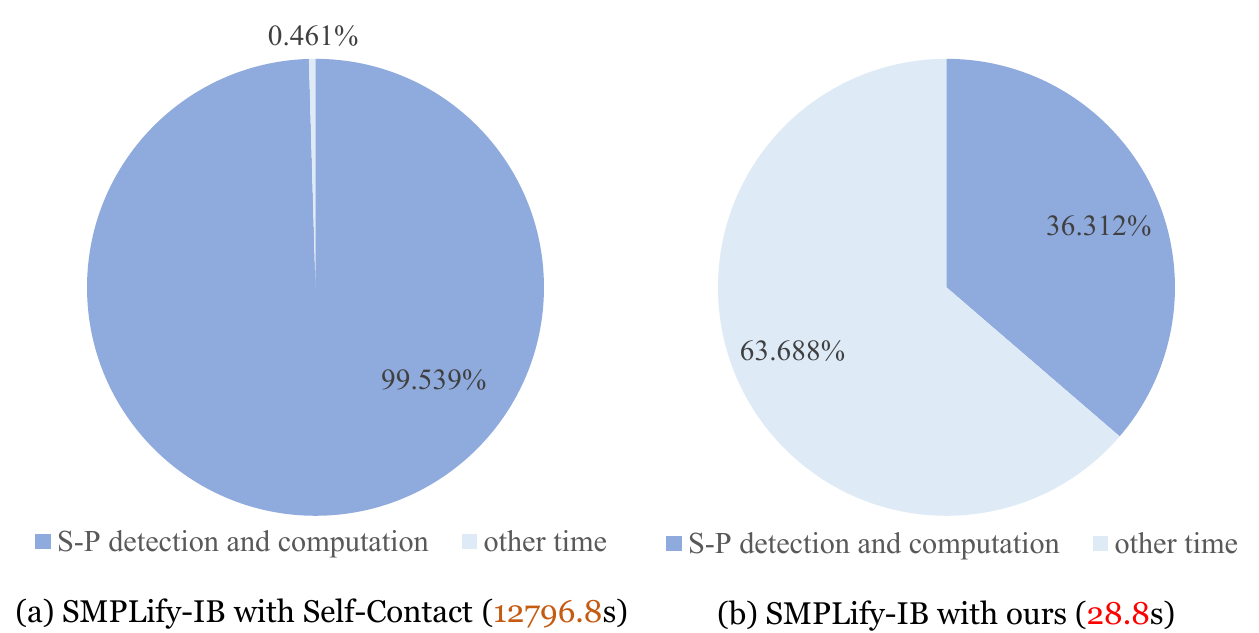}
  \caption{\textbf{Time consumption when deploying the two self-penetration detection and computation algorithms in our optimization routine.} We count the time taken in an optimization stage with 500 iterations on a single batch (128 frames) and document the proportion of time spent by the self-penetration modules in the overall duration~(in deep blue).} 
  \label{fig: sup_smplify_time}
\end{figure}
\cref{fig: sup_smplify_time} provides quantitative comparisons on time consumption of our optimization routines with SOTA self-penetration algorithm~(Self-Contact in SMPLify-XMC~\cite{muller2021self}) and our proposed light-weight approach~(downsample 1/3 version). The experiment is conducted on a NVIDIA. 3090 GPU, with each optimization performing with 500 iterations on a single batch (128 frames). While the Self-Contact algorithm yields high detection accuracy, it comes at a significant time and computational expense~(\ie, nearly 100s per frame on a RTX3090 GPU). Our detection module brings nearly 450 times speed while archiving comparable self-penetration refinement.

\subsection{Visualizations for gravity-based constraints.}
\begin{figure}[t]
  \centering
  \includegraphics[width=\linewidth]{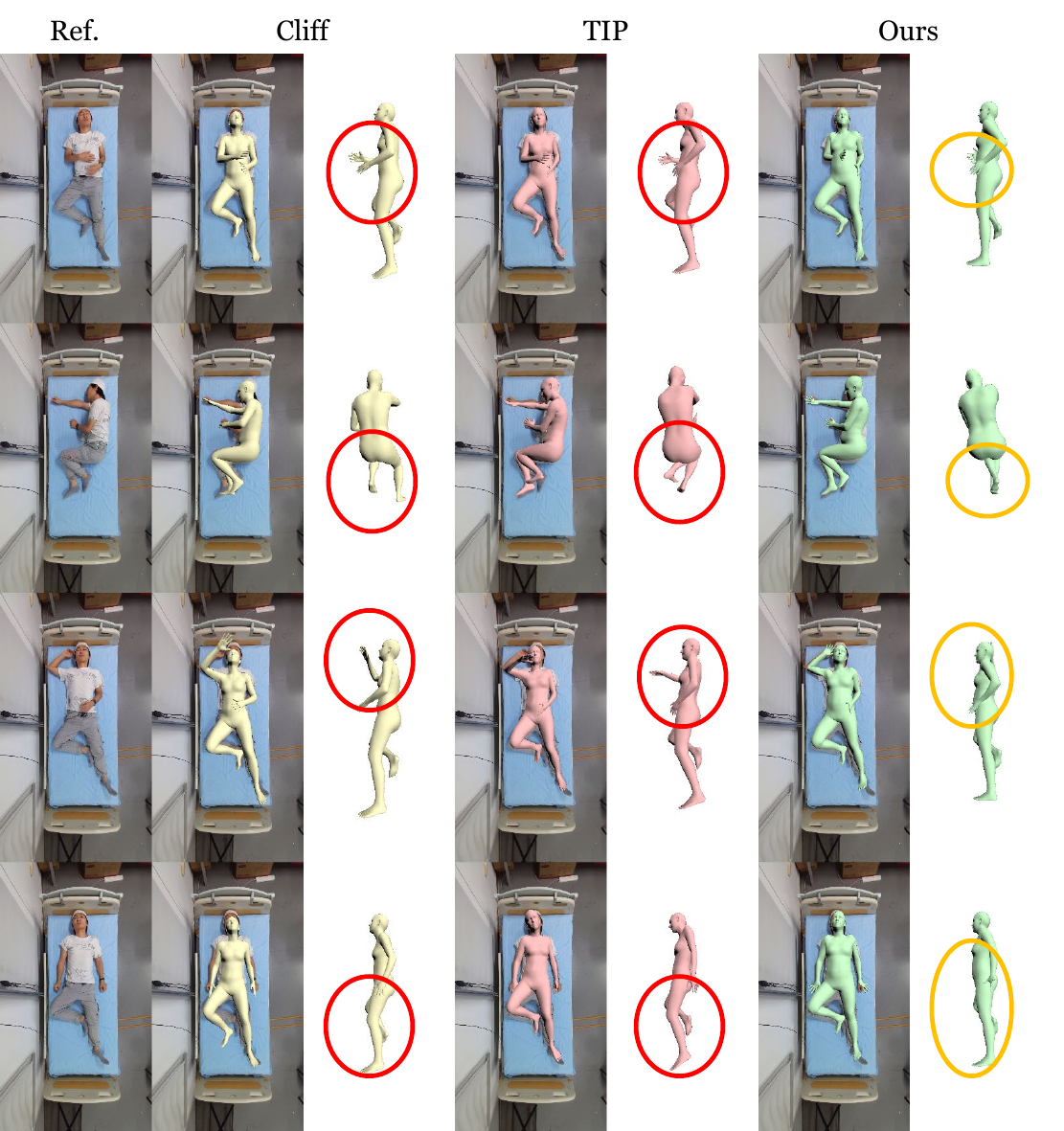}
  \caption{\textbf{Qualitative comparisons on the p-GTs generated by Cliff~(predicted on images), TIP and our generations by SMPLify-IB.} We highlight the implausible limb lifts by single-view depth ambiguity in red ellipses and our refinement with yellow ellipses.} 
  \label{fig: sup_smplify_gravity}
\end{figure}

~\cref{fig: sup_smplify_gravity} provides more visual evidence on the efficiency of our gravity constraints in SMPLify-IB. Traditional single-view regression-based method~(yellow meshes by Cliff) and optimization-based method~(red meshes by a SMPLify-like approach adopted in TIP) face serious depth ambiguity in the in-bed scene, especially when limbs overlap from the camera perspective, thus leading to implausible limb lifts~(\eg, hand lifts in the first and third rows in~\cref{fig: sup_smplify_gravity}, and leg lifts when legs contact and overlap in the third row). Our proposed gravity constraints, accompanied by a strong self-penetration detection and penalty term, effectively alleviate the depth ambiguity issue while maintaining reasonable contact. This validates the feasibility of alleviating depth ambiguity issues with physical constraints in specific scenarios.

\subsection{Failure cases for SMPLify-IB}

\begin{figure}[t]
  \centering
  \includegraphics[width=\linewidth]{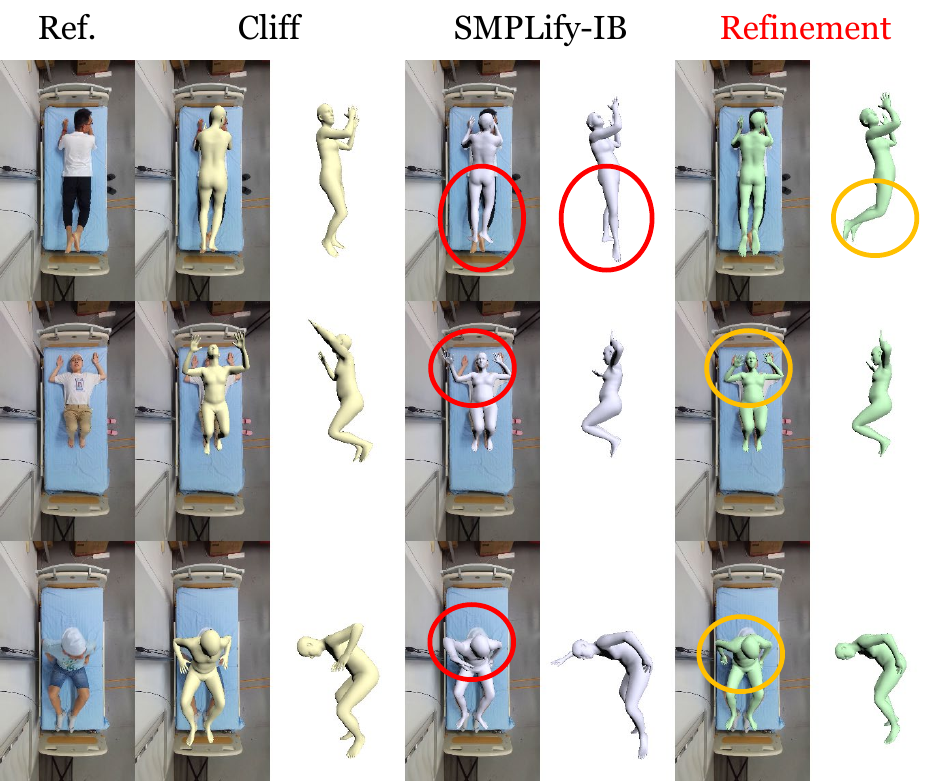}
  \caption{\textbf{Typical failure cases of SMPLify-IB.} We highlight the wrong generations with red markers and our refinement in the yellow ellipses.} 
  \label{fig: sup_smplify_failure_case}
\end{figure}

About 1.6\% samples of our optimization results might fail due to severely false initialization by CLIFF, wrong judgment in gravity constraints, and trade-offs in the multiple-term optimization, as presented in~\cref{fig: sup_smplify_failure_case}. Thus we manually inspected all generated results and carried out another round of optimization to address these errors, aiming at generating reliable p-GTs for the TIP dataset. The refinement is highlighted with yellow ellipses in~\cref{fig: sup_smplify_failure_case}. 

\subsection{Failure cases for PI-HMR}
\begin{figure}[htbp]
  \centering
  \includegraphics[width=\linewidth]{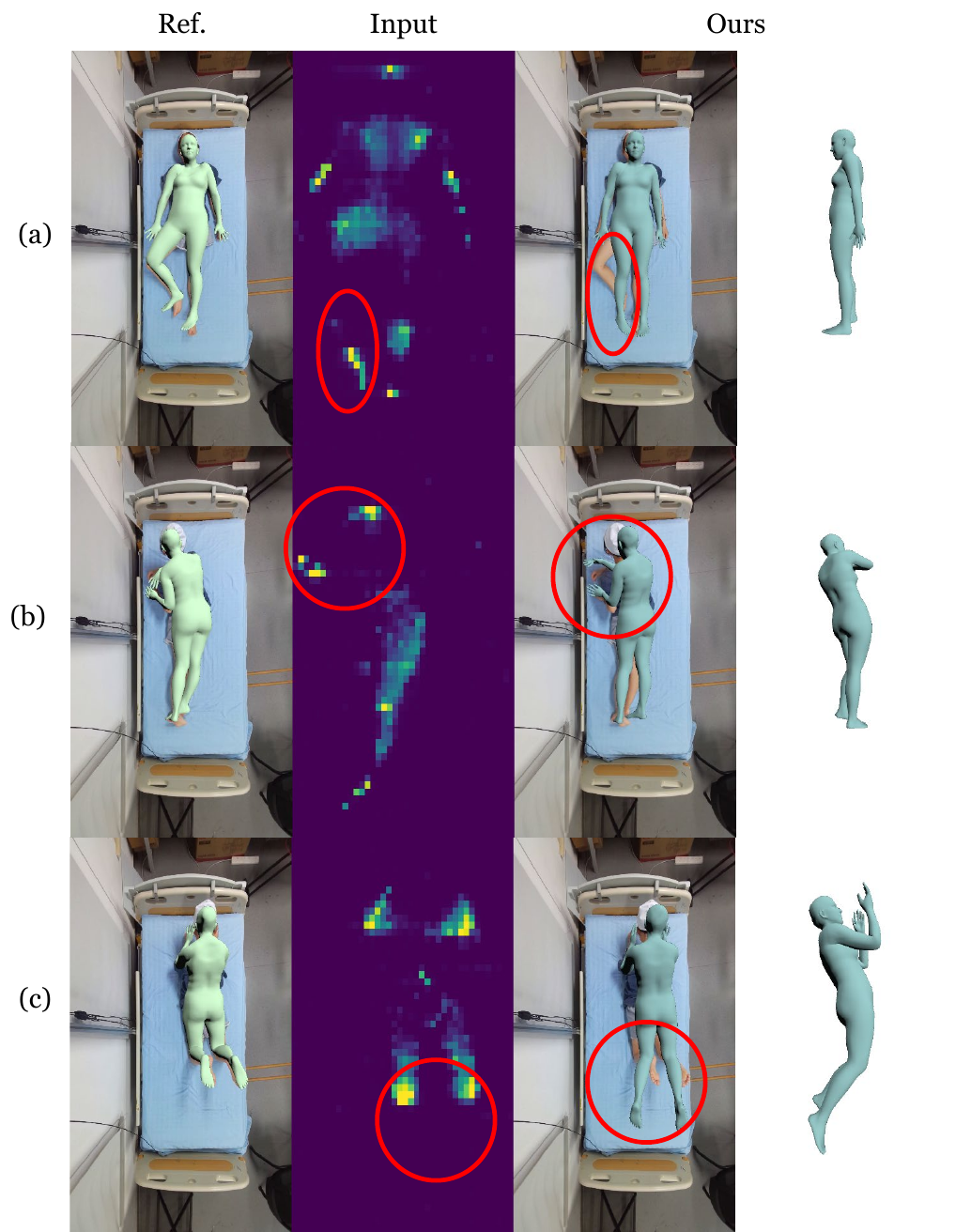}
  \caption{\textbf{Typical failure cases of PI-HMR.} We highlight the mispredictions and corresponding pressure regions with red markers.} 
  \label{fig: sup_pihmr_failure_case}
\end{figure}

In~\cref{fig: sup_pihmr_failure_case}, we show a few examples where PI-HMR fails to reconstruct reasonable human bodies. The reason mainly falls in the information ambiguities, ranging from (a) PI-HMR mistakenly identifies the contact pressure between the foot and the bed as originating from the leg~(shown in the red ellipse), (b) hand lifts and (c) leg lifts.

\subsection{More Qualitative Visualizations}

\begin{figure*}[htbp]
  \centering
  \includegraphics[width=\linewidth]{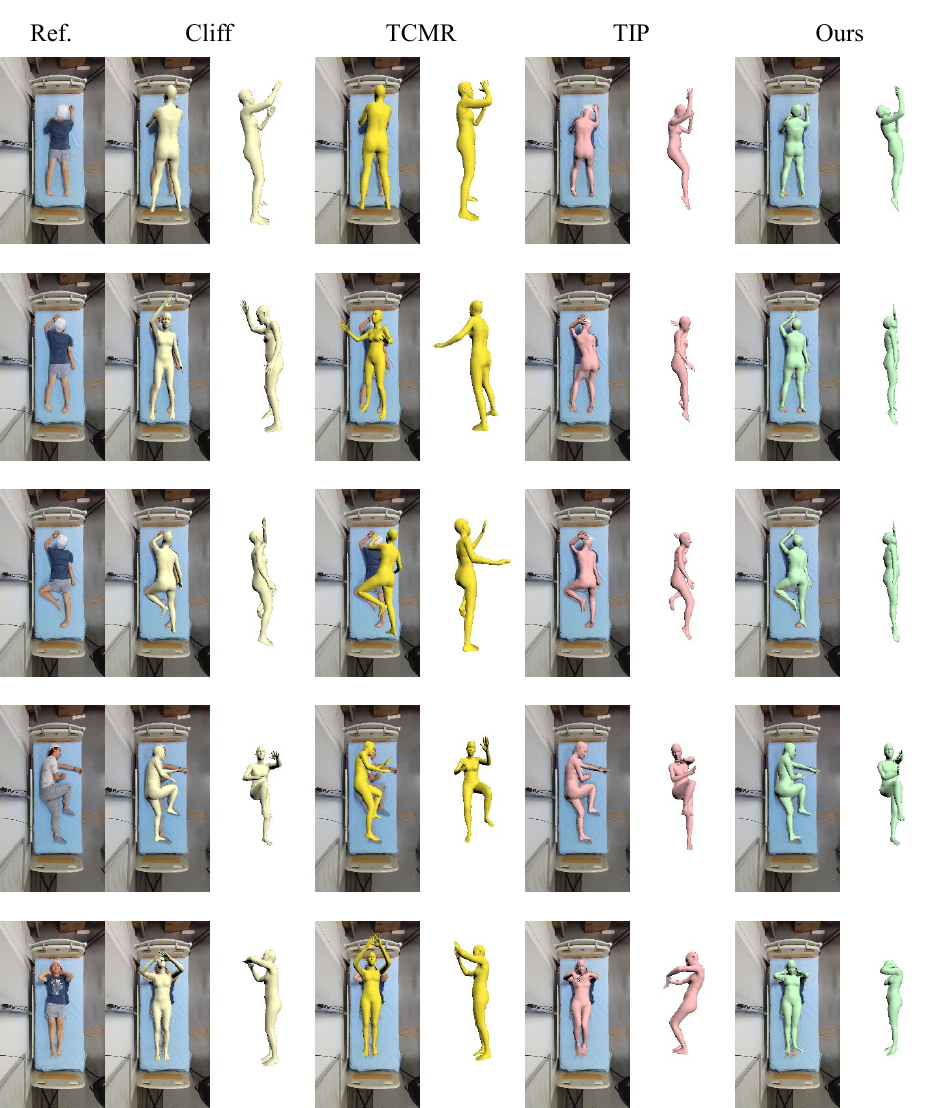}
  \caption{\textbf{Qualitative results of our generated p-GTs on the TIP dataset.} We compare our results with SOTA vision-based methods Cliff and TCMR~(predicted on RGB images) and p-GTs provided in TIP.} 
  \label{fig: sup_simplify_quali}
\end{figure*}

\begin{figure*}[htbp]
  \centering
  \includegraphics[width=\linewidth]{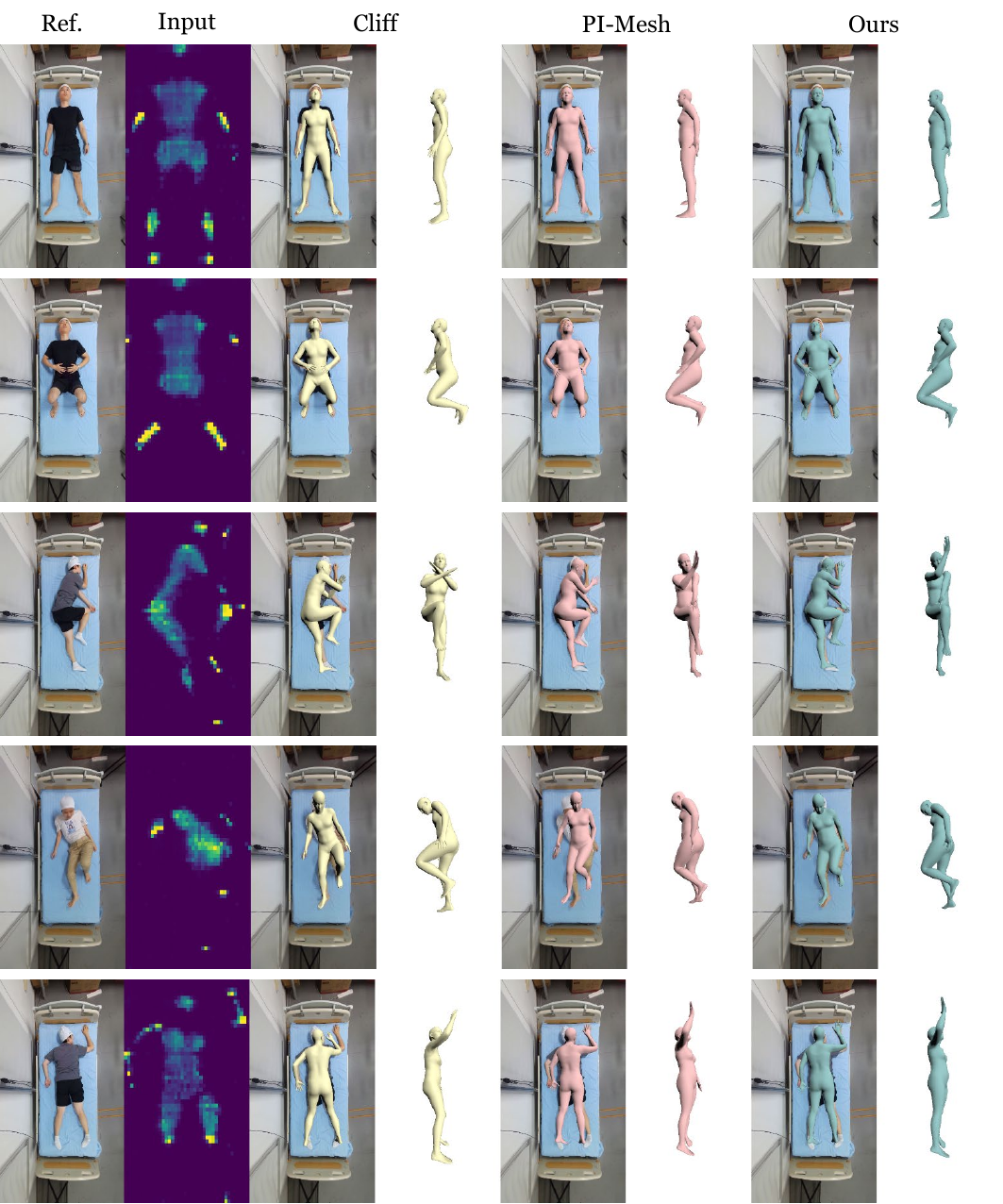}
  \caption{\textbf{Qualitative results of PI-HMR's performance on the TIP dataset.} We compare our results with SOTA vision-based methods Cliff~(predicted on RGB images) and pressure-based method PI-Mesh.} 
  \label{fig: sup_pihmr_quali}
\end{figure*}

\begin{figure*}[htbp]
  \centering
  \includegraphics[width=\linewidth]{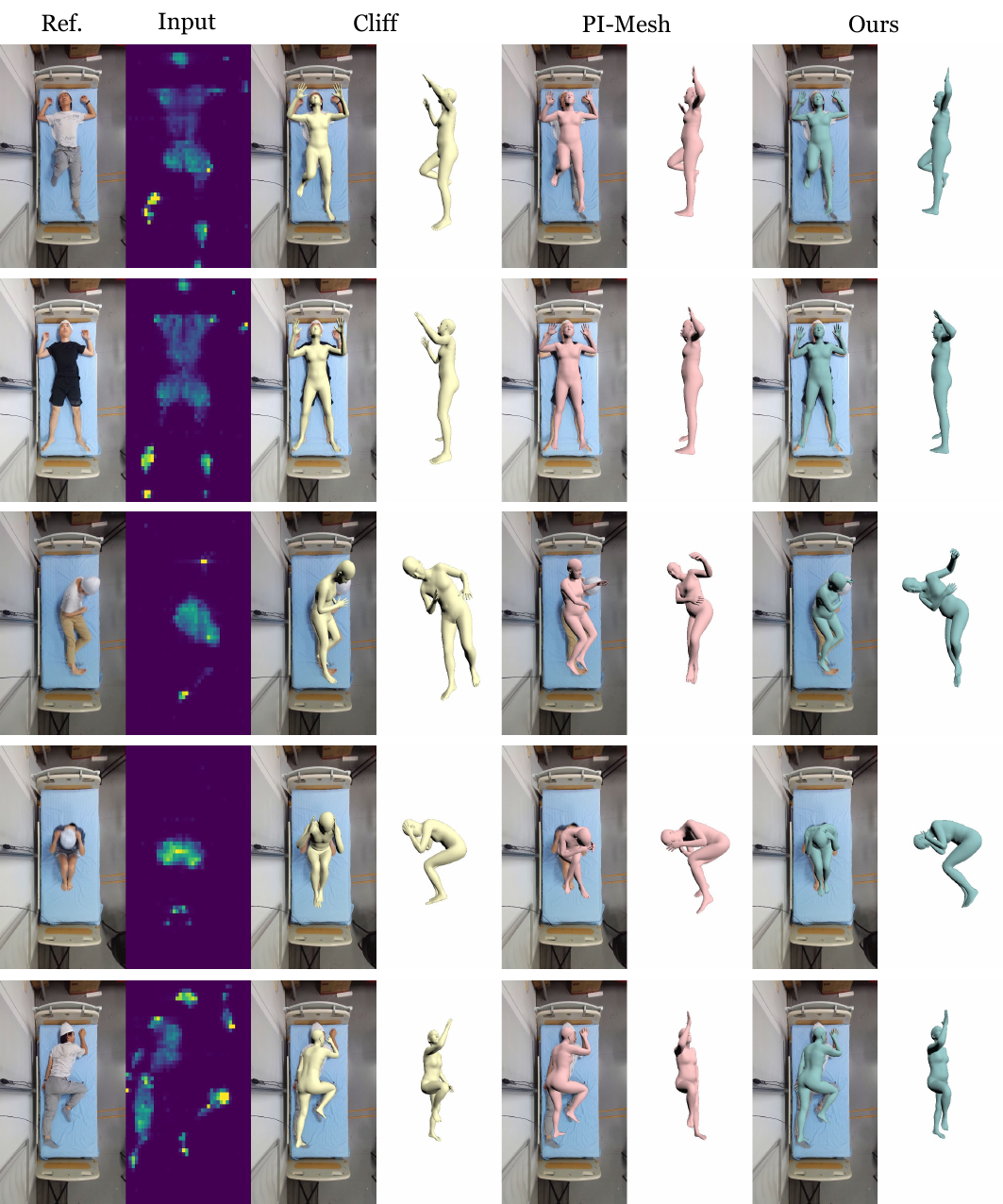}
  \caption{\textbf{More qualitative results of PI-HMR's performance on the TIP dataset.}} 
  \label{fig: sup_pihmr_quali_2}
\end{figure*}

We present more qualitative visualizations on the performance of our proposed optimization strategy SMPLify-IB in~\cref{fig: sup_simplify_quali} and PI-HMR in~\cref{fig: sup_pihmr_quali} and~\cref{fig: sup_pihmr_quali_2}.

\section{Limitations and Future works}

we conclude our limitations and future works in three main aspects:

(1) \textbf{Hand and foot parametric representations:} More diverse and flexible tactile interactions exist in the in-bed scenarios. For instance, the poses of the hands and feet vary with different human postures, thereby influencing the patterns of localized pressure. However, the SMPL model fails to accurately depict the poses of hands and feet, thereby calling for more fine-grained parametric body representations~\cite{pavlakos2019expressive, osman2022supr} to precisely delineate the contact patterns between human bodies and the environment.

(2) \textbf{Explicit constraints from contact cues:} In this work, we propose an end-to-end learning approach to predict human motions directly from pressure data. The learning-based pipeline can rapidly sense the pressure distribution patterns and generate high-quality predictions from pressure sequences, yet it may lead to underutilization of contact priors from pressure sensors and cause misalignment between limb position and contact regions~(\eg, torso and limbs lift). In future works, we aim to explicitly incorporate contact priors through learning or optimization methods~\cite{shimada2023decaf} to further enhance the authenticity of the model's predictions.

(3) \textbf{Efforts for information ambiguity:} In this work, we aspire to mitigate the information ambiguity issue through pressure-based feature sampling and habit-based Test-Time Optimization strategies, yielding accuracy improvement; however, challenges persist. Building upon the observation that users perform movements in certain habitual patterns, we expect to develop a larger-scale motion generation model reliant on VQ-VAE~\cite{van2017neural} or diffusion~\cite{ho2020denoising} techniques, to address the deficiencies in single-pressure modality based on users' motion patterns.

\newpage

\newpage

\newpage

\clearpage
{
\small
\noindent
\textbf{Acknowledgements:} We thank the anonymous reviewers for their suggestions. This work is supported by the National Natural Science Foundation of China under Grant No. 62072420.
}

{
    \small
    \bibliographystyle{ieeenat_fullname}
    \bibliography{main}
}

\end{document}